\documentclass{article}

\PassOptionsToPackage{numbers, compress}{natbib}

\usepackage[final]{neurips_2022}

\usepackage[utf8]{inputenc} % allow utf-8 input
\usepackage[T1]{fontenc}    % use 8-bit T1 fonts
\usepackage{nicefrac}       % compact symbols for 1/2, etc.
\usepackage{microtype}      % microtypography
\usepackage{threeparttable}
\usepackage{zshi}

\title{Efficiently Computing Local Lipschitz Constants of Neural Networks via Bound Propagation}

\author{\normalsize
Zhouxing Shi\textsuperscript{1},
\enskip Yihan Wang\textsuperscript{1},
\enskip Huan Zhang\textsuperscript{2},
\enskip Zico Kolter\textsuperscript{2,3},
\enskip Cho-Jui Hsieh\textsuperscript{1}\\
{\normalsize \textsuperscript{1}University of California, Los Angeles \enskip \textsuperscript{2}Carnegie Mellon University \enskip
\textsuperscript{3}Bosch Center for AI}\\
{\tt\small zshi@cs.ucla.edu, yihanwang@cs.ucla.edu, huan@huan-zhang.com}\\
{\tt\small zkolter@cs.cmu.edu, chohsieh@cs.ucla.edu}\\ 
}

\newcommand{\rebuttal}[1]{{#1}}

\begin{document}

\maketitle

\begin{abstract}
Lipschitz constants are connected to many properties  of neural networks, such as robustness, fairness, and generalization. Existing methods for computing Lipschitz constants either produce relatively loose upper bounds or are limited to small networks. In this paper, we develop an efficient framework for computing the $\ell_\infty$ local Lipschitz constant of a neural network by tightly upper bounding the norm of Clarke Jacobian via linear bound propagation. We formulate the computation of local Lipschitz constants with a linear bound propagation process on a high-order backward graph induced by the chain rule of Clarke Jacobian. To enable linear bound propagation, we derive tight linear relaxations for specific nonlinearities in Clarke Jacobian. This formulate unifies existing ad-hoc approaches such as RecurJac, which can be seen as a special case of ours with weaker relaxations. The bound propagation framework also allows us to easily borrow the popular Branch-and-Bound (BaB) approach from neural network verification to further tighten Lipschitz constants. Experiments show that on tiny models, our method produces comparable bounds compared to exact methods that cannot scale to slightly larger models; on larger models, our method efficiently produces tighter results than existing relaxed or naive methods, and our method scales to much larger practical models that previous works could not handle. We also demonstrate an application on provable monotonicity analysis. Code is available at \url{https://github.com/shizhouxing/Local-Lipschitz-Constants}.

\end{abstract}

\section{Introduction}

Lipschitz constants are important for characterizing many properties of neural networks, including robustness~\citep{szegedy2013intriguing,hein2017formal,tsuzuku2018lipschitz,zhang2019recurjac,huang2021training}, fairness~\citep{dwork2012fairness,kang2020inform}, generalization~\citep{bartlett2017spectrally}, and explanation~\citep{fel2022good}. Local Lipschitz constants, which only need to hold for a small local region, can more precisely characterize the local behavior of a network. Intuitively they characterize how fast the output of the network changes between any two input within the region.

It is challenging to exactly and efficiently compute local Lipschitz constants. Naive approaches such as computing the product of the induced norm for all layers cannot capture local information and typically produce vacuous bounds. To compute tight and exact local Lipschitz constants, \citet{jordan2020exactly} considered small ReLU networks (e.g., up to tens of neurons), by solving a  mixed-integer programming (MIP) problem to bound the norm of Clarke's generalized Jacobian~\citep{clarke1975generalized}. However, solving MIP is often too costly and cannot scale to slightly larger networks. On the other hand, RecurJac~\citep{zhang2019recurjac} (an improved version of Fast-Lip~\citep{weng2018towards}) is a specialized recursive algorithm for bounding the Jacobian and computing local Lipschitz constants, which is also relatively efficient but the produced bounds are relatively loose. Moreover, most of the existing works on local Lipschitz constants only used small toy models and cannot feasibly handle larger practical networks.

Recently, in the field of neural network verification, many methods are proposed to compute provable output bounds for neural networks~\citep{katz2017reluplex,bunel2018unified,dvijotham2018dual}; especially, linear bound propagation methods~\citep{wong2018provable,wang2018formal,zhang2018efficient,singh2019abstract} are becoming very successful~\citep{wang2021beta,zhang2022gcpcrown} because they are scalable and can be efficiently accelerated on GPUs. These methods propagate the linear relationship between layers, where nonlinearities in networks are relaxed into linear bounds. In this work, we ask the question if we can borrow these successful techniques from neural network verification to scale up the computation of local Lipschitz constants to larger and more practical networks.
%In this work, we aim to bound the Clarke Jacobian of a neural network rather than the output of the network itself, which has not been handled by existing linear bound propagation works. 

In this paper, we aim to efficiently compute relatively tight $\ell_\infty$ local Lipschitz constants for neural networks using the bound propagation framework. We formulate this problem as upper bounding the $\ell_\infty$ norm of the Clarke Jacobian, and we formulate the computation for the Clarke Jacobian from a chain rule and its norm as a \emph{higher-order backward computational graph} augmented to the original forward graph of the network.
On the augmented computational graph, we generalize linear bound propagation to bound the Clarke Jacobian, and we thereby reformulate the problem of computing local Lipschitz constants under a linear bound propagation framework. On the backward graph, applying Clarke gradients in the chain rule is nonlinear and requires a linear relaxation. It is essentially formed by a \emph{group} of functions and is different from \emph{single} activation functions in regular neural network verification. We propose a tight and closed-form linear relaxation for Clarke gradients with an optimality guarantee on the tightness, and thereby we efficiently bound the Clarke Jacobian with linear bound propagation.
We also show that RecurJac is a special case under our formulation where loose interval bounds instead of tight linear relaxation are used for nontrivial cases.
Our formulation also allows us to develop a scalable and flexible framework enhanced by progress from recent neural network verifiers using linear bound propagation. We demonstrate that we can further tighten our bounds by Branch-and-Bound when time budget allows, for a trade-off between tightness and time cost.

Experiments show that our method efficiently produces tightest $\ell_\infty$ local Lipschitz constants compared to other relaxed methods, and is much more efficient than the exact MIP method. Moreover, our method scales to much larger models including practical convolutional neural networks (CNN) that previous works could not handle. We also demonstrate an application of our method for provably analyzing the monotonicity of neural networks.

\section{Related Work}

For neural networks, a loose global Lipschitz constant (upper bound) can be computed by the product of layer-wise induced norms~\citep{szegedy2013intriguing}. Based on this product, \citet{virmaux2018lipschitz} considered the effect of activation and approximated upper bounds which, however, are not guaranteed; \citet{gouk2021regularisation} used a power method for convolutional layers. LipSDP~\citep{fazlyab2019efficient} used semidefinite programming (SDP) to compute tighter bounds. These works all compute global Lipschitz constants which can be much looser than local Lipschitz constants as they have to hold even for distinct input points and they cannot characterize the local behavior of neural networks. 

We focus on local Lipschitz constants in this paper.
On local Lipschitz constants,
LipMIP~\citep{jordan2020exactly} used mixed integer programming (MIP) to compute exact results.
LipOpt~\citep{latorre2019lipschitz} used polynomial optimization but is limited to smooth activations not including the widely used ReLU.
LipBaB~\citep{bhowmick2021lipbab} combined relatively loose interval bound propagation~\citep{hickey2001interval,mirman2018differentiable,gowal2018effectiveness} with branch-and-bound to compute exact local Lipschitz constants.
All of these methods cannot scale to relatively larger models due to their computational cost.
FastLip~\citep{weng2018towards} and its improved version RecurJac~\citep{zhang2019recurjac} used recursive procedures to bound the Jacobian. While FastLip and RecurJac are much more efficient, their bounds are relatively loose due to the use of strictly looser relaxations compared to ours.
In contrast, we compute $\ell_\infty$ local Lipschitz constants efficiently while our bounds are tighter than existing relaxed methods.
Besides, there are also several other works on Lipschitz constants under different settings:
\citet{avant2021analytical} derived layer-wise analytical bounds but only on a simplified definition for Lipschitz constants, which does not cover Lipschitzness in the entire local region;
and \citet{laurel2022dual} proposed a dual number abstraction method to bound the Clarke Jacobian but they focused on non-smooth perturbations that can be represented by a single scalar, while we consider high dimensional perturbations.

Lipschitz constants can also be used to train certifiably robust neural networks, by computing margins from Lipschitz constants~\citep{tsuzuku2018lipschitz,leino2021globally,huang2021training} or enforcing 1-Lipschitzness~\citep{anil2019sorting,li2019preventing,singla2021skew,zhang2021towards,zhang2021boosting,zhang2022rethinking}, 
These methods are competitive compared to certified training by directly bounding the output~\citep{mirman2018differentiable,gowal2018effectiveness,zhang2019towards,shi2021fast,wang2021convergence}, but they are beyond the scope of this paper as we focus on pre-trained neural networks.
\section{Background}

\subsection{ReLU Network}

Suppose $f(\rvx)$ is a $K$-way neural network classifier given a $d$-dimensional input $\rvx\in\sR^d$, and then  $f(\rvx)\in\sR^K$. For the simplicity of presentation, we mainly focus on feedforward ReLU networks, but our method can also be applied to general network architectures and activations as will be discussed in Section~\ref{sec:method}. 
Suppose the network has $n$ layers, it takes input $h_0(\rvx)=\rvx$ and then computes 
$$ \forall i\in[n],\,z_i(\rvx)= \rmW_i h_{i-1}(\rvx) + \rvb_i,$$ 
$$ \forall i\in[n-1],\,h_i(\rvx)=\sigma(z_i(\rvx)),\,h_n(\rvx)=z_n(\rvx), $$
where $i\in[n]$ means $1\leq i\leq n$, $ z_i(\rvx) $ is the pre-activation output of the $i$-th linear layer with weight $\rmW_i$ and bias $\rvb_i$, and $ h_i(\rvx)$ is the output after activation $\sigma(\cdot)$. This formulation is compatible with CNNs since convolutional layers are also linear.

\subsection{Lipschiz Constant}
The $(\alpha,\beta)$-Lipschitz constant of a network $f(\rvx)$ over an open set $\gX\in\sR^d$ is defined as:
\begin{equation}
 L^{(\alpha,\beta)}(f,\gX)=\sup_{\rvx_1, \rvx_2 \in \gX, \enskip \rvx_1\neq \rvx_2 } \frac{\|f(\rvx_1)-f(\rvx_2)\|_\beta}{\|\rvx_1-\rvx_2\|_\alpha}.
 \label{eq:def_lip}  
\end{equation} 
If $f$ is smooth and $(\alpha,\beta)$-Lipschitz continuous over $\gX$, the Lipschitz constant can be computed by upper bounding the norm of Jacobian, i.e., $L^{(\alpha,\beta)}(f,\gX)=\sup_{\rvx\in\gX} \|\nabla f(\rvx)\|_{\alpha,\beta}$.
Since neural networks with non-smooth ReLU activation are non-smooth functions, we consider Clarke Jacobian~\citep{clarke1975generalized} instead, which is defined as the convex hull of $\lim_{i\rightarrow\infty} \nabla f(\rvx_i)$ for any sequence $\{\rvx_i\}_{i=1}^\infty$ such that every $f(\rvx_i)$ is differentiable at $\rvx_i$ respectively. We denote the Clarke Jacobian at $\rvx$ as $\partial f(\rvx)$, and then
\begin{equation}
L^{(\alpha,\beta)}(f,\gX) = \sup_{\rvx\in\gX,~ \rmJ(\rvx)\in\partial f(\rvx)} \|\rmJ(\rvx)\|_{\alpha,\beta}.
\label{eq:lip_def}
\end{equation}

Global Lipschitz constant, which considers the supremum over $\gX=\sR^d$, needs to guarantee \eqref{eq:def_lip} even for distant $\rvx_1,\rvx_2$, and thus it can be loose and cannot capture the local behavior of the network.
We focus on $\ell_\infty$ local Lipschitz constants, where $\gX=B_\infty(\rvx_0,\eps)\coloneqq\{\rvx:\|\rvx-\rvx_0\|_\infty\leq \eps \} $ is a small $\ell_\infty$-ball with radius $\eps$ around $\rvx_0$, and we take $\alpha=\beta=\infty$ for the definition in \eqref{eq:def_lip}.
While exactly computing \eqref{eq:lip_def} is possible for very small networks~\citep{jordan2020exactly,bhowmick2021lipbab}, for slightly larger networks, we only expect to compute guaranteed upper bounds for \eqref{eq:def_lip} and \eqref{eq:lip_def}, and we aim to make the upper bounds as tight as possible with an acceptable computational cost.

\subsection{Backward Linear Bound Propagation}
\label{sec:bound_propagation}
To bound the Clarke Jacobian, we will use linear bound propagation which is originally used for certifiably bounding the output of neural networks in neural network verification. We adopt backward bound propagation~\citep{zhang2018efficient,wong2018provable,singh2019abstract,xu2020automatic} which typically propagates the linear relationship between the output layer to be bounded and all the previous layers it depends on, in a backward manner.
Suppose we want to compute certified bounds for output layer $ h_n(\rvx)$ w.r.t. all $\rvx$ from a small domain $ B_\infty(\rvx_0,\eps)$. Starting from the output layer, we take $ \rmA_n=\rmI$, and then $ h_n(\rvx) = \rmA_n h_n(\rvx) $. For every $ i \in [n]$, suppose $ h_i(\rvx) $ can be bounded by linear functions w.r.t. $h_{i-1}(\rvx)$ parameterized by $\ul{\rmP}_i, \ol{\rmP}_i, \ul{\rvq}_i, \ol{\rvq}_i$ (the bounds element-wisely hold):
\begin{equation}
\ul{\rmP}_i h_{i-1}(\rvx) + \ul{\rvq}_i \leq h_i(\rvx) \leq \ol{\rmP}_i h_{i-1}(\rvx) + \ol{\rvq}_i.
\label{eq:linear_relaxation}
\end{equation}
Then for every $ i=n,n-1,\cdots,1$ in order, $ \rmA_i h_i(\rvx) $ can be recursively bounded by substituting $h_i(\rvx)$ with \eqref{eq:linear_relaxation}:
\begin{align}
\rmA_i h_i(\rvx) \leq \rmA_{i-1} h_{i-1}(\rvx) + \rvc_i, 
\quad
\rmA_{i-1} = [\rmA_i]_+\ol{\rmP}_i + [\rmA_i]_- \ul{\rmP}_i, ~\rvc_i = [\rmA_i]_+\ol{\rvq}_i + [\rmA_i]_-\ul{\rvq}_i,
\label{eq:apply_relaxation}
\end{align}
where $[\cdot]_+$ stands for taking positive elements from the matrix or vector, and vice versa for $[\cdot]_-$. $\rmA_{i-1}$ stands for the coefficients  of the linear relationship propagated from layer $i$ to layer $i\!-\!1$, and $\rvc_i$ is a bias term produced at this layer. By this recursive procedure in a backward manner and accumulating $ \rvc_i (i\in[n])$, eventually bounds are propagated to the input layer as $h_n(\rvx)\leq \rmA_0 \rvx + \sum_{i=1}^n \rvc_i$, which directly represents the linear relationship between the output layer and $\rvx$. 
We can eliminate $\rvx$ from $\rmA_0\rvx$ to obtain the final bounds. When $ \rvx\!\in\! B_\infty(\rvx_0,\eps)$, for the $j$-th dimension of $h_n(\rvx)$, we can use the fact that $ [\rmA_0]_{j,:}\rvx \leq [\rmA_0]_{j,:} \rvx_0 + \eps\|[\rmA_0]_{j,:}\|_{1}$, where $ [\rmA_0]_{j,:} $ stands for the $j$-th row in $\rmA_0$.

Parameters in \eqref{eq:linear_relaxation} can be obtained by considering the relation between $ h_i(\rvx)$ and $h_{i-1}(\rvx)$.   We can first obtain bounds  $ \rvl_i\!\leq\!  z_i(\rvx) \!\leq\!\rvu_i $ using bound propagation by viewing $z_i(\rvx)$ as the output layer. Then a linear relaxation bounds activation $h_i(\rvx)=\sigma(z_i(\rvx))$ with linear functions of $z_i(\rvx)$:
\begin{equation}
\forall \rvl_i \leq z_i(\rvx)\leq \rvu_i, \enskip 
\ul{\rvs}_i z_i(\rvx) + \ul{\rvt}_i \leq  \sigma(z_i(\rvx)) \leq \ol{\rvs}_i z_i(\rvx) + \ol{\rvt}_i.
\label{eq:relu_relaxation}
\end{equation}
Intuitively, for the $j$-th neuron, $[\ul{\rvs}_i]_j [z_i(\rvx)]_j + [\ul{\rvt}_i]_j$ and $[\ol{\rvs}_i]_j [z_i(\rvx)]_j + [\ol{\rvt}_i]_j$ are two lines bounding $ \sigma([z_i(\rvx)]_j)$ against input $[z_i(\rvx)]_j$.
We provide a detailed derivation and illustration for ReLU in Appendix~\ref{ap:relu}.
Then to satisfy  \eqref{eq:linear_relaxation}, we can take: 
\begin{align}
\ul{\rmP}_i = \diag(\ul{\rvs}_i) \rmW_{i}, 
\enskip \ol{\rmP}_i = \diag(\ol{\rvs}_i) \rmW_{i},
\enskip \ul{\rvq}_i = \diag(\ul{\rvs}_i)\rvb_{i} + \ul{\rvt}_i,
\enskip \ol{\rvq}_i = \diag(\ol{\rvs}_i)\rvb_{i} + \ol{\rvt}_i.\label{eq:pq}
\end{align}
Interested readers can find more details of backward bound propagation in the literature~\citep{zhang2018efficient,wang2021beta}. In this work, we generalize this linear bound propagation  to bound the Clarke Jacobian.

\section{Methodology}
\label{sec:method}

\begin{figure*}[t]
\centering
\includegraphics[width=\textwidth]{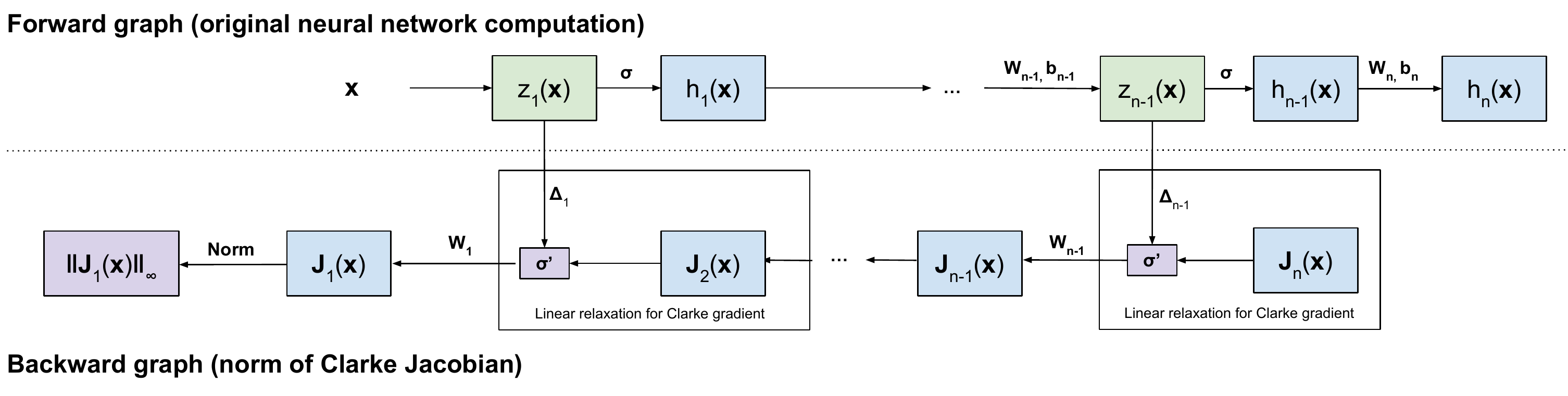}
\caption{Illustration of our proposed framework where we formulate the computation for local Lipschitz constants by augmenting the original computation graph of the neural network (forward graph) with an additional subgraph for computing the norm of Clarke Jacobian (backward graph). The backward graph relies on the pre-activation values ($\rvz_1(\rvx),\cdots,\rvz_{n-1}(\rvx)$) in the forward graph to compute the Clarke gradient of activations.
Linear relaxation for two categories of nonlinearities in the backward graph (nodes in purple), norm ($ \|\rmJ_1(\rvx)\|_\infty$) and Clarke gradient ($ \sigma' $), are detailed in Figure~\ref{fig:abs}, Figure~\ref{fig:clarke} and Figure~\ref{fig:clarke_subgraph}.
We also perform branch-and-bound (BaB) to tighten the bounds by branching pre-activation nodes in the forward graph (nodes in green).}
\label{fig:framework}
\end{figure*}

We present our method in this section. We first formulate the local Lipschitz constant computation as a higher-order backward computational graph, and use bound propagation to obtain the bounds of the Clarke Jacobian. We discuss our methods for tightly bounding nonlinearities in Clarke Jacobian, and we show that RecurJac~\citep{zhang2019recurjac} is a special case with looser interval bounds. Finally, we use Branch-and-Bound to further tighten the bounds.

\subsection{Clarke Jacobian for ReLU Networks}
\label{sec:clarke_jacobian}

In ReLU networks, the ReLU activation for a single neuron is defined as $\sigma(x)=\max\{x,0\}$ which is non-smooth at $x\!=\!0$. We consider its Clarke gradient~\citep{clarke1975generalized} denoted as $\partial\sigma(x)$: when $x\!<\!0$, $\partial \sigma(x)\!=\!\{0\}$; when $x\!>\!0 $, $ \partial\sigma(x)\!=\!\{1\}$; and when $x\!=\!0$, $\partial\sigma(x)\!=\![0,1]$.
For the $i$-th layer $(1\!\leq\! i\!<\!n)$, we use $\partial\sigma([z_i(\rvx)]_j) $ to denote the Clarke gradient of the $j$-th neuron and $\partial\sigma(z_i(\rvx))$ to denote the Clarke gradient for all neurons in the layer.
And we use a diagonal matrix $\Delta_i(\rvx)\in \partial \sigma(z_i(\rvx))$ to denote a Clarke gradient of neurons in the $i$-th layer layer, where $[\Delta_i(\rvx)]_{jj}\in \partial\sigma([z_i(\rvx) ]_j)$.
Then a chain rule can be used to compute the Clarke Jacobian for the whole network~\citep{imbert2002support,jordan2020exactly}. Denote $\rmJ_i(\rvx) \in \frac{\partial f(\rvx)}{\partial h_{i-1}(\rvx)}$ as a Clarke Jacobian w.r.t. the $(i\!-\!1)$-th layer in the chain rule. We have $\rmJ_n(\rvx)=\rmW_n$, 
and for all $ i\in[n-1]$, we have  
\begin{align}
\rmJ_i(\rvx)\in 
\frac{\partial f(\rvx)}{\partial h_{i-1}(\rvx)}
\Big\{ \rmJ_{i+1}(\rvx) \Delta_i(\rvx) \rmW_i: \rmJ_{i+1}(\rvx) \in \frac{\partial f(\rvx)}{\partial h_{i}(\rvx)}
,\, \Delta_i(\rvx) \in \partial\sigma(z_i(\rvx))
\Big\}.
\label{eq:jacobian}
\end{align}
Finally $\rmJ_1(\rvx)\in\partial f(\rvx)$ is a Clarke Jacobian of the entire network~\citep{jordan2020exactly}.
Since $f(\rvx)\in\sR^K$ and $x\in\sR^d$, we have $ \rmJ_1(\rvx)\in\sR^{K\times d}$. 
As we aim to compute the $\ell_\infty$ local Lipschitz constant, we upper bound  $\|\rmJ_1(\rvx)\|_\infty~(\rmJ_1(\rvx)\in \partial f(x))$ w.r.t. all $\rvx\in\gX$.

Given the forward pass computation of the original network, we can formulate a computational graph for $\|\rmJ_1(\rvx)\|_\infty$ in a backward pass accordingly as shown in Figure~\ref{fig:framework}. We augment the original forward graph with a backward graph, where the backward graph has a dependency on the pre-activation outputs of layers in the forward graph, due to $ \Delta_1,\Delta_2,\cdots,\Delta_{n-1}$. We build the augmented graph by traversing the forward graph, similar to automatic differentiation for neural network training. The formulation with a computational graph allows our method to easily generalize to different network architectures given the existing bound propagation framework for general neural networks~\citep{xu2020automatic}.

We use the bound propagation framework described in Section~\ref{sec:bound_propagation} to upper bound $\|\rmJ_1(\rvx)\|_\infty$.
On the backward graph, when bounds are propagated to $\rmJ_i(\rvx)$, similar to $\rmA_i$ and $\rvc_i$ in \eqref{eq:apply_relaxation},
we use $ \widetilde{\rmA}_i $ to denote the coefficient matrix and $ \widetilde{\rvc}_i$ to denote the bias term for the backward graph, and then we have:
\begin{equation}
\forall i\in[n],\enskip
\| \rmJ_1(\rvx) \|_\infty \leq 
\rmJ_i(\rvx) \widetilde{\rmA}_i + \sum_{j=0}^{i-1} \widetilde{\rvc}_j.
\label{eq:backward_bound_propagation}  
\end{equation}
Note that we have $\widetilde{\rmA}_i$ on the right of $\rmJ_i(\rvx)$ rather than the left, since $\Delta_i(\rvx)\rmW_i$ is multiplied on the right of $ \rmJ_{i+1}(\rvx)$ in \eqref{eq:jacobian}.
As intermediate bounds for $z_i(\rvx)$ are needed for bounding the forward graph, we also need intermediate bounds $\rmL_i\!\leq\!\rmJ_i(\rvx)\!\leq\!\rmU_i~(\forall \rvx\!\in\!\gX)$ for the backward graph by starting bound propagation from $\rmJ_i(\rvx)$ instead of $\|\rmJ_1(\rvx)\|_\infty$. These intermediate bounds are used in the linear relaxation for nonlinearities in the backward graph.
There are particularly two categories of nonlinearities to be tackled on the backward graph:
1) $\ell_\infty$ norm in $\|\rmJ_1(\rvx)\|_{\infty}$; 
2) $ \rmJ_{i+1}(\rvx)\Delta_i~(\forall i\!\in\![n-1])$ in \eqref{eq:jacobian}, which applies the Clarke gradient of ReLU on the later layer's Clarke Jacobian.  
We handle them in Section~\ref{sec:norm} and Section~\ref{sec:grad_relax} respectively.
Then bounds can be propagated starting at $ \| \rmJ_1(\rvx)\|_{\infty}$, to $ \rmJ_1(\rvx), \rmJ_2(\rvx),\cdots,\rmJ_n(\rvx)$ in order. 
Bounds are eventually propagated to $ \rmJ_n(\rvx)$ as \eqref{eq:backward_bound_propagation} with $i=n$, where $\rmJ_n(\rvx)$ can be substituted with $\rmW_n$, and thereby we obtain a final upper bound for $\|\rmJ_1(\rvx)\|_\infty$. 

\subsection{Norm of Clarke Jacobian}
\label{sec:norm}

The $\ell_\infty$ norm of Clarke Jacobian is the first nonlinearity we would encounter during backward bound propagation, as it is the last node in the backward graph (see Figure~\ref{fig:framework}).
The norm is computed as 
$\|\rmJ_1(\rvx)\|_\infty = \max_{1\leq k\leq K} \sum_{j=1}^d \big|[\rmJ_1(\rvx) ]_{kj}\big|$, 
where $K$ rows in $\rmJ_1(\rvx)$ can be bounded separately, and after that we can aggregate results on $K$ rows by taking the $\max$.
For the simplicity of the following analysis, we assume $K=1$ as we can handle one row in $ \rmJ_1(\rvx)$ at each time, but they can still be batched in implementation. 
Then Clarke Jacobian $\rmJ_i(\rvx)~(i\in[n])$ and its bounds $\rmL_i,\,\rmU_i$ can be viewed as row vectors, and we will use $[\cdot]_j$ to denote the $j$-th element. 

By a tight linear relaxation for $ |\cdot|$ as shown in Figure~\ref{fig:abs} in Appendix~\ref{ap:abs} (only the upper bound is needed), we have the following bound for $  \|\rmJ_1(\rvx)\|_\infty$:
\begin{proposition}[Linear relaxations for matrix $\| \cdot \|_\infty$ norm]
For all $x\in\gX$, suppose $\forall j\!\in\![d]$, $[\rmL_1]_j \!\leq\! [\rmJ_1(\rvx)]_j\!\leq\! [\rmU_1]_j$, and $\rmJ_1(\rvx)$ is a row vector, we have a bound for its matrix $\|\cdot\|_\infty$ norm:
\begin{align*}
&\|\rmJ_1(\rvx)\|_\infty \leq \rmJ_1(\rvx)\widetilde{\rmA}_1  + \widetilde{\rvc}_0,\\
& \text{where}\enskip  [\widetilde{\rmA}_1]_j=
\left\{\begin{aligned}
\frac{\big|[\rmU_1]_j\big|-\big|[\rmL_1]_j\big|}{[\rmU_1]_j-[\rmL_1]_j} & \quad [\rmL_1]_j<[\rmU_1]_j\\
0 & \quad [\rmL_1]_j=[\rmU_1]_j
\end{aligned}
\right.,
\quad [\widetilde{\rvc}_0]_j=-[\widetilde{\rmA}_1]_j[\rmL_1]_j+\big|[\rmL_1]_j\big|.
\end{align*} 
\label{prop:abs_norm}
\end{proposition}
We prove it in Appendix~\ref{ap:abs_norm}, and this propagates bounds from the norm of Clarke Jacobian to $ \rmJ_1(\rvx)$. 

\subsection{Clarke Gradient of Activation Functions}
\label{sec:bound_jacobian}
\label{sec:grad_relax}

For each layer $i\in[n-1]$ in the chain rule for computing the Clarke Jacobian as \eqref{eq:jacobian}, in addition to the fixed weight matrix $\rmW_i$, there is a $\rmJ_{i+1}(\rvx) \Delta_i(\rvx)$ term, where $\Delta_i(\rvx)$ is determined by pre-activation value $z_i(\rvx)$ from the forward graph. 
For the $j$-th neuron in the layer, 
$[\rmJ_{i+1}(\rvx) \Delta_i(\rvx)]_j
= [\rmJ_{i+1}(\rvx)]_j [\Delta_i(\rvx)]_{jj}$.
When $ [\rvu_i]_j<0 $ or $[\rvl_i]_j>0$, we have $[\Delta_i(\rvx)]_{jj}\!=\!0$ and $[\Delta_i(\rvx)]_{jj}\!=\!1$ respectively fixed, and then for these special cases $ [\rmJ_{i+1}(\rvx)\Delta_i(\rvx)]_j$ is already linear w.r.t. $ [\rmJ_{i+1}(\rvx)]_j$, and thus a linear relaxation is not needed. 
Otherwise, $[\rmJ_{i+1}(\rvx) \Delta_i(\rvx)]_j$ is nonlinear, since both $[\rmJ_{i+1}(\rvx)]_j$ and $[\Delta(\rvx)]_{jj}$ may vary given different $\rvx\in\gX$.
We thereby relax the nonlinearity $[\rmJ_{i+1}(\rvx)]_j[\Delta_i(\rvx)]_{jj}$ for $ [\Delta_i(\rvx)]_{jj}=[0,1]$. 

\begin{figure}[ht]
\begin{minipage}{.48\textwidth}
\centering
\includegraphics[width=.90\textwidth]{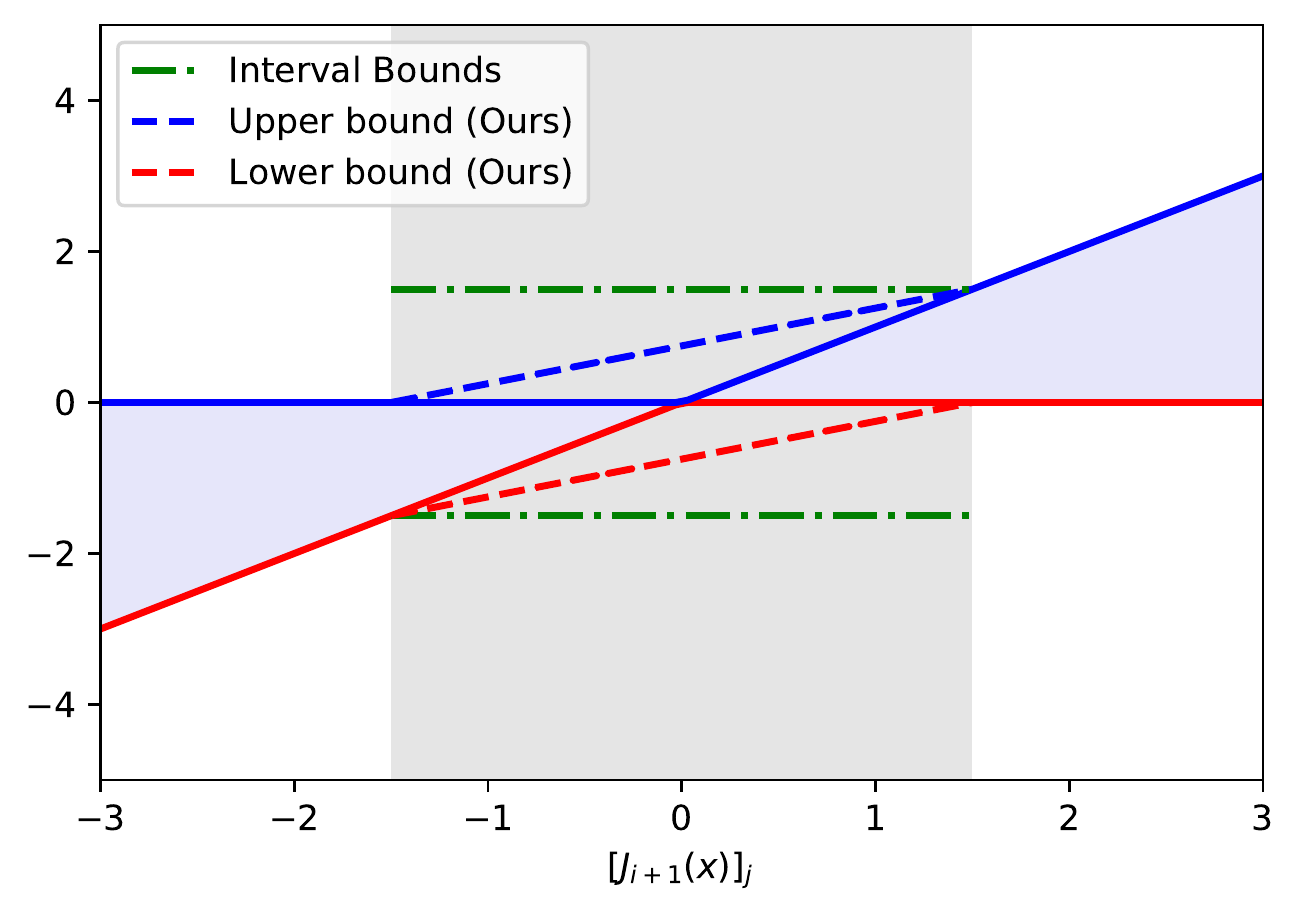}
\caption{For a neuron $j$ in layer $i$, when $[\Delta_i(\rvx)]_{jj}\!=\![0,1]$, we plot $ [\rmJ_{i+1}(\rvx)\Delta_i(\rvx)]_j $ as a group of functions w.r.t. $ [\rmJ_{i+1}(\rvx)]_j$ (violet area). In this example, $ -1.5\!\leq\! [\rmJ_{i+1}(\rvx)]_j\!\leq\!1.5$. Our linear relaxation (dashed blue and green lines) is much tighter than interval bounds (dashed green lines) by \citet{zhang2019recurjac}.}
\label{fig:clarke}
\end{minipage}
\hspace{15pt}
\begin{minipage}{.45\textwidth}
\centering
\includegraphics[width=.98\textwidth]{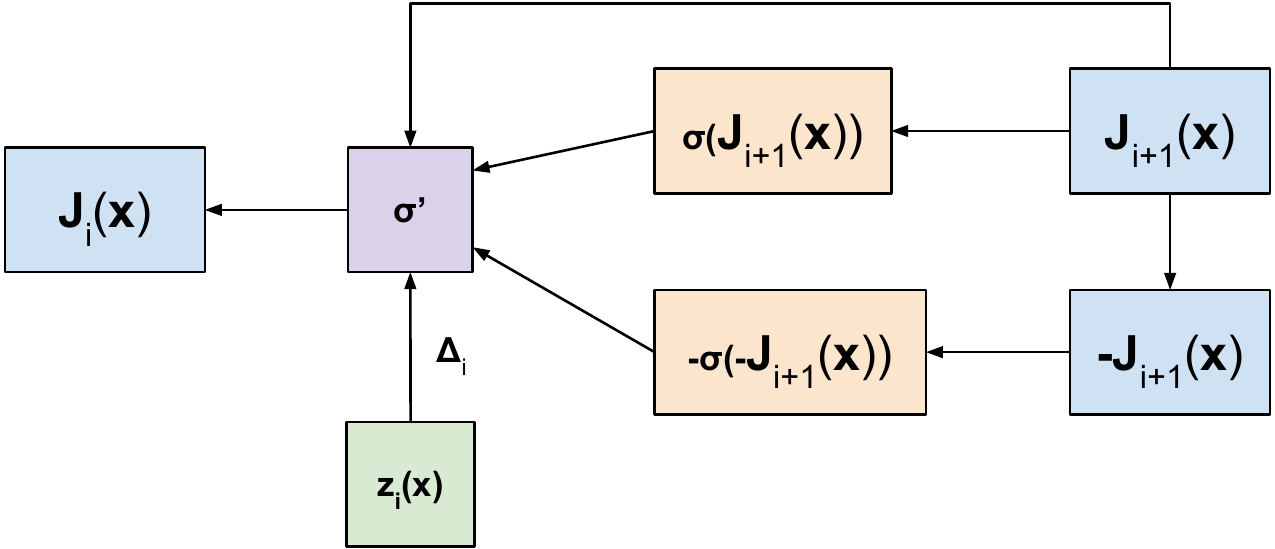}
\caption{Computation graph for bounding each Clarke gradient. It is the internal computation for boxes in Figure~\ref{fig:framework}. Two ReLU nodes are inserted as per Proposition~\ref{prop:grad_bound}. In bound propagation, the $\sigma'$ node takes intermediate bounds from $z_i(\rvx)$ to decide whether bounds are directly propagated from $\rmJ_{i+1}(\rvx)$ (no relaxation needed), or indirectly via $ \sigma(\rmJ_{i+1}(\rvx))$ and $-\sigma(-\rmJ_{i+1}(\rvx))$ (with relaxation), for each neuron respectively. }
\label{fig:clarke_subgraph}
\end{minipage}

\end{figure}

As illustrated in Figure~\ref{fig:clarke}, the solid blue line and solid red line are the exact upper and lower bound respectively for $[\rmJ_{i+1}(\rvx) \Delta_i(\rvx)]_{j}$, which are piecewise linear w.r.t. $[\rmJ_{i+1}(\rvx)]_j$.
When $ [\rmL_{i+1}]_j\geq 0$ or $[\rmU_{i+1}]_j\leq 0$, $[\rmJ_{i+1}(\rvx) \Delta_i(\rvx)]_{j}$ can be exactly bounded by linear functions w.r.t. $[\rmJ_{i+1}(\rvx)]_j$:
\begin{align}
\left\{\enskip
\begin{aligned}
0
\leq [\rmJ_{i+1}(\rvx) \Delta_i(\rvx)]_{j}
\leq [\rmJ_{i+1}(\rvx)]_j
& \quad \text{when}~[\rmL_{i+1}]_j\geq 0,[\Delta_i(\rvx)]_{jj}=[0,1],\\
[\rmJ_{i+1}(\rvx)
\leq [\rmJ_{i+1}(\rvx) \Delta_i(\rvx)]_{j}
\leq 0
& \quad \text{when}~[\rmU_{i+1}]_j\leq 0,[\Delta_i(\rvx)]_{jj}=[0,1].
\end{aligned}
\right.
\label{eq:trivial_relaxation}
\end{align}

The most nontrivial case appears when $ [\rmL_{i+1}]_j\!<\!0\!<\![\rmU_{i+1}]_j$, where exact bounds for $[\rmJ_{i+1}(\rvx) \Delta_i(\rvx)]_{j}$ are nonlinear given the input region.
We relax the exact bounds into linear bounds that can be handled by linear bound propagation.
We observe that the exact upper bound is essentially a ReLU function of $\rmJ_{i+1}(\rvx)$ (solid blue line in  Figure~\ref{fig:clarke}), and the exact lower bound can be viewed as a ReLU function flipped both horizontally and vertically (solid red line in Figure~\ref{fig:clarke}). 
Formally, we have:
\begin{proposition}[Exact lower and and upper bounds for the Clarke gradient of a ReLU neuron]
For the $j$-th neuron in layer $i(i\!\in\![n-1])$, if $[\Delta_i(\rvx)]_{jj}=[0,1]$, we have:
\begin{align}
[ \rmJ_{i+1}(\rvx)\Delta_i(\rvx)]_j \geq   \min\{ [\rmJ_{i+1}(\rvx)]_j, 0 \},
\enskip [ \rmJ_{i+1}(\rvx)\Delta_i(\rvx)]_j \leq \max\{ [\rmJ_{i+1}(\rvx)]_j, 0\},\label{eq:jacobian_max}
\end{align}
which can be rewritten using ReLU activation:
\begin{align}
& \min\{ [\rmJ_{i+1}(\rvx)]_j, 0 \} = -\sigma( - [\rmJ_{i+1}(\rvx)]_j ), 
\enskip \max\{ [\rmJ_{i+1}(\rvx)]_j, 0\} = \sigma([\rmJ_{i+1}(\rvx)]_j)
\label{eq:jacobian_relu}. 
\end{align}
\label{prop:grad_bound}
\end{proposition}
\vspace{-15pt}
Its correctness can be easily verified by considering the sign of $[\rmJ_{i+1}(\rvx)]_j$ as shown in Appendix~\ref{ap:grad_bound}.
We then decompose $ [\rmJ_{i+1}(\rvx)\Delta_i(\rvx)]_j$ into two ReLU activations (\eqref{eq:jacobian_relu}), as illustrated in Figure~\ref{fig:clarke_subgraph}.

ReLU can be further relaxed by \eqref{eq:relu_relaxation}, as derived in Appendix~\ref{ap:relax_clarke_grad}, which yields:
\begin{proposition}[Linear relaxations for Clarke Jacobian of a ReLU neuron]
For the $j$-th neuron in the $i$-th layer ($i\!\in\![n-1]$), given $[\Delta_i(\rvx)]_{jj}=[0,1]$ and $[\rmL_{i+1}]_j\leq [\rmJ_{i+1}(\rvx)]_j\leq [\rmU_{i+1}]_j$,
we have the following relaxation:
\begin{align*}
[\rmJ_{i+1}(\rvx)\Delta_i(\rvx)]_j \geq 
[\widetilde{\ul{\rvs}}_{i+1}]_j [\rmJ_{i+1}(\rvx)]_j + [\widetilde{\ul{\rvt}}_{i+1}]_j,
\enskip [\rmJ_{i+1}(\rvx)\Delta_i(\rvx)]_j \leq [\widetilde{\ol{\rvs}}_{i+1}]_j [\rmJ_{i+1}(\rvx)]_j + [\widetilde{\ol{\rvt}}_{i+1}]_j,
\end{align*}
\begin{align*}
\text{where}\quad& 
[\widetilde{\ul{\rvs}}_{i+1}]_j= \frac{ -\sigma(-[\rmU_{i+1}]_j) + \sigma(-[\rmL_{i+1}]_j) }{ [\rmU_{i+1}]_j - [\rmL_{i+1}]_j },
\enskip [\widetilde{\ul{\rvt}}_{i+1}]_j= - [\widetilde{\ul{\rvs}}_{i+1}]_j [\rmL_{i+1}]_j -\sigma(-[\rmL_{i+1}]_j),\\
&[\widetilde{\ol{\rvs}}_{i+1}]_j= \frac{ \sigma([\rmU_{i+1}]_j) - \sigma([\rmL_{i+1}]_j) }{ [\rmU_{i+1}]_j - [\rmL_{i+1}]_j },
\enskip [\widetilde{\ol{\rvt}}_{i+1}]_j= - [\widetilde{\ol{\rvs}}_{i+1}]_j [\rmL_{i+1}]_j + \sigma([\rmL_{i+1}]_j).
\end{align*}
\label{prop:relax_clarke_grad}
\end{proposition}
\vspace{-10pt}

The linear relaxation in Proposition~\ref{prop:relax_clarke_grad} corresponds to dashed blue and red lines shown in Figure~\ref{fig:clarke} for the relaxed upper bound and lower bound respectively.
And \eqref{eq:trivial_relaxation} is a special case for the relaxation in Proposition~\ref{prop:relax_clarke_grad}.
In the following theorem, we show that relaxation in Proposition~\ref{prop:relax_clarke_grad} is the tightest linear relaxation:

\begin{theorem}[Optimality of linear relaxations for the Clarke gradient of a ReLU neuron]
For the $j$-th neuron in the $i$-th layer ($i\in[n-1]$), 
and any linear bound coefficients $ \ul{s},\ul{t},\ol{s},\ol{t}$ satisfying 
\begin{equation}
\forall J\!\in\! \Big[[\rmL_{i+1}]_j, [\rmU_{i+1}]_j\Big], \forall [\Delta_i(\rvx)]_{jj}\!\in\![0,1],
\enskip \ul{s}J+\ul{t} \leq J \cdot [\Delta_i(\rvx)]_{jj} \leq \ol{s}J+\ol{t}, 
\label{eq:clarke_other_relaxation}
\end{equation}
we show that they produce looser relaxations:
\begin{align*}
&(\ul{s},\ul{t})\neq ([\widetilde{\ul{\rvs}}_{i+1}]_j,[\widetilde{\ul{\rvt}}_{i+1}]_j) \implies \forall [\rmL_{i+1}]_j\!<\!J\!<\![\rmU_{i+1}]_j,\enskip\ul{s}J+\ul{t}<[\widetilde{\ul{\rvs}}_{i+1}]_jJ+[\widetilde{\ul{\rvt}}_{i+1}]_j,\\
&(\ol{s},\ol{t})\neq ([\widetilde{\ol{\rvs}}_{i+1}]_j,[\widetilde{\ol{\rvt}}_{i+1}]_j) \implies \forall [\rmL_{i+1}]_j\!<\!J\!<\![\rmU_{i+1}]_j,\enskip\ol{s}J+\ol{t}>[\widetilde{\ol{\rvs}}_{i+1}]_jJ+[\widetilde{\ol{\rvt}}_{i+1}]_j,
\end{align*}
where $[\widetilde{\ul{\rvs}}_{i+1}]_j,[\widetilde{\ul{\rvt}}_{i+1}]_j,[\widetilde{\ol{\rvs}}_{i+1}]_j,[\widetilde{\ol{\rvt}}_{i+1}]_j$ are defined in Proposition~\ref{prop:relax_clarke_grad}. 
\label{theorem:tightness}
\end{theorem}
This theorem, proved in Appendix~\ref{ap:tightness} states that our proposed linear relaxation is provably tighter than any other valid linear relaxation.
While there are previous works tightening the linear relaxation for activation functions~\citep{zhang2018efficient,lyu2019fastened,xu2020fast}, we consider the optimal linear relaxation for a group of functions (as shown in Figure~\ref{fig:clarke}) instead of a single activation function, and our relaxation is closed-form with a provable optimality, without gradient-based optimization~\citep{lyu2019fastened,xu2020fast}.

\subsection{Connections to RecurJac~\citep{zhang2019recurjac}}

While RecurJac is a recursive algorithm specialized for local Lipschitz constants and it was not related to bound propagation in the original paper, we find that it is essentially a special case under our formulation with linear bound propagation, but it uses looser interval bound relaxation instead of our tight linear relaxation proposed in Proposition~\ref{prop:grad_bound} and Proposition~\ref{prop:relax_clarke_grad} (note that \emph{interval} bound propagation is also a special case of \emph{linear} bound propagation, but interval bounds use relaxation with zero slope only instead of general non-zero slopes).
Figure~\ref{fig:clarke} shows a nontrivial case where $[\rmL_{i+1}]_j<0<[\rmU_{i+1}]_j$ and $[\Delta_i(\rvx)]_{jj}=[0,1]$. RecurJac takes interval bounds as the horizontal green dashed lines with a zero slope, and the interval bounds are looser than our linear relaxation with non-zero slopes, i.e., the gap between upper and lower bounds is larger.
In Appendix~\ref{ap:recurjac}, we show that if we use interval relaxation when $[\rmL_{i+1}]_j<0<[\rmU_{i+1}]_j$, our framework will be equivalent to RecurJac. 
Compared to RecurJac, we not only have a more general formulation with linear bound propagation but also produce tighter results.

\subsection{Branch and Bound}

Formulating the backward computational graph also allows us to utilize recent progress in linear bound propagation for neural network verification, to further tighten the results. In neural network verification, Branch-and-Bound (BaB) has been used to compute tighter bounds~\citep{katz2017reluplex,bunel2018unified,bunel2020branch,wang2018efficient,wang2018formal,xu2020fast,wang2021beta} by branching activations or input and then bounding each smaller subdomain respectively. We also utilize BaB to tighten our bounds when time budget allows.
 
We denote $\gC_0=\{(\rvl_1, \rvu_1), (\rvl_2, \rvu_2), \cdots(\rvl_{n-1}, \rvu_{n-1}) \} $ as the domain of all pre-activation bounds of ReLU neurons in the forward computational graphs before BaB. We use $\gC$ to denote a pool of domains that we currently have, and initially $ \gC=\{ \gC_0 \}$.  BaB aims to recursively split domains in $\gC $ into smaller subdomains $\gC_1, \gC_2,\cdots,$ where $\gC_0 = \gC_1 \bigcup \gC_2 \bigcup \cdots$, and bounds for each subdomain can be computed respectively using bound propagation, as linear relaxation can generally be tighter with smaller domains.

At each iteration, we take $B$ domains in $\gC$ with loosest bounds, where $B$ is the batch size for BaB. 
For each domain, we choose a layer $ \hat{i} $ and a ReLU neuron $\hat{j} $ in layer $\hat{i}$ on the forward graph, such that the neuron has uncertain Clarke gradient ($ [\rvl_{\hat{i}}]_{\hat{j}}\!\leq\! 0\!\leq\! [\rvu_{\hat{i}}]_{\hat{j}} $).
We branch this domain into two subdomains by branching $ ([\rvl_{\hat{i}}]_{\hat{j}}, [\rvu_{\hat{i}}]_{\hat{j}})$ into $ ([\rvl_{\hat{i}}]_{\hat{j}}, -\tilde{\eps}) $ and $ (\tilde{\eps}, [\rvl_{\hat{i}}]_{\hat{j}}) $ respectively, where $\tilde{\eps}$ is a sufficiently small value (e.g., $10^{-9}$), so that the Clarke gradient for the branched neuron becomes certain in each subdomain (0 and 1 respectively) with other neurons unchanged. We compute the bounds for the new subdomains, and we repeat this process until reaching a time limit or there is no domain left to branch. Finally, the bound of the domain with the loosest bound in $\gC$ is the tightened result.

We use a heuristic score to decide which neuron to branch for a given domain, by estimating the potential improvement to the bounds~\citep{bunel2018unified}.
For branching a ReLU neuron on the forward graph, we estimate the potential improvement on the backward graph. Suppose neuron $\hat{j}$ in layer $\hat{i}$ is branched, we estimate the gap between the lower and upper bound in the liner relaxation, as the gap can be closed after the branching. We estimate the gap (the gap between the blue dashed line and the red dashed line in Figure~\ref{fig:clarke}) as 
$ \frac{1}{2} ( [\rmU_{\hat{i}+1}]_{\hat{j}} -  [\rmL_{\hat{i}+1}]_{\hat{j}} )^2$, and we multiply it by coefficient $[\widetilde{\rmA}_{\hat{i}}]_{\hat{j}} $ from the bound propagation as the heuristic score. We thereby branch neurons with highest scores.

\section{Experiments}

In the experiments, we focus on evaluating the tightness of \rebuttal{$\ell_\infty$ local Lipschitz constant bounds} and the computational cost. 
There are three parts: 
1) We compare the tightness and efficiency on relatively small models and synthetic data, to accommodate slower baselines;
2) We also use practical image datasets with larger models;
3) Finally, we demonstrate an application of our method on analyzing the monotonicity of neural networks.
Additional experimental details are provided in Appendix~\ref{apd:exp_details}.

\paragraph{Baselines}
We compare our methods to the following baselines:
\textbf{NaiveUB} multiplies the induced norm of each layer, which can scale to arbitrarily large models but the bound is often vacuous.
\textbf{LipMIP}~\cite{jordan2020exactly} computes exact and tightest local Lipschitz constants using MIP but can only work for tiny models. During the process of solving MIP, an upper bound of the result may be available and is gradually improved, and thus for models that LipMIP cannot finish within a time budget, we report the upper bound obtained at the timeout.
\textbf{LipSDP}~\cite{fazlyab2019efficient} computes upper bounds of $\ell_2$ local Lipschitz constants using SDP, and we convert their $\ell_2$ results into upper bounds for the $\ell_\infty$-case by a $ \sqrt{d}$ factor.
Since LipSDP by itself does not support the $\ell_\infty$ case which may not be the intended use of LipSDP, the results converted from the $\ell_2$ case are for reference only. 
\textbf{LipBaB}~\citep{bhowmick2021lipbab} uses loose interval bounds with branch-and-bound.
\textbf{RecurJac}~\citep{zhang2019recurjac} is a recursive algorithm computing relaxed local Lipschitz constants which can be seen as a special case of ours with weaker relaxations.
The baselines except NaiveUB do not support CNN in their implementation, and we convert the CNN models to equivalent MLP models for these baselines.

\vspace{-10pt}
\begin{table}[ht]
\centering
\caption{
Local Lipschitz constant values and runtime (seconds) on MLP and CNN models with growing width for a 16-dimensional synthetic input data point. Smaller values are tighter results. Width for MLP stands for number of neurons in each hidden layer, and width for CNN stands for number of filters in each convolutional layer. 
``C'' and ``F'' in the model names for CNNs denote the number of convolutional layers and fully-connected layers respectively.
We set a timeout of 1000s for LipMIP and LipSDP, and 60s for BaB.
``*'' denotes that we report the upper bound LipMIP returns at timeout, and ``-'' means LipSDP cannot return an upper bound at timeout.
}  
\adjustbox{max width=.995\textwidth}{
    \begin{tabular}{c|cccccc|cccccc}
    \toprule
    \multirow{3}{*}{Method} & \multicolumn{6}{c|}{3-Layer MLP} & \multicolumn{6}{c}{CNN-2C1F}\\
    & \multicolumn{2}{c}{Width=32} & \multicolumn{2}{c}{Width=64} & \multicolumn{2}{c|}{Width=128} & \multicolumn{2}{c}{Width=4} & \multicolumn{2}{c}{Width=8} & \multicolumn{2}{c}{Width=16}\\
    & Value & Runtime &  Value & Runtime &  Value & Runtime &  Value & Runtime &  Value & Runtime &  Value & Runtime \\
    \midrule
    
    NaiveUB & 24.31	& 0.05& 33.02&	0.01&	49.39&	0.00
   & 24.38&	0.00&	39.99&	0.01&	51.68&	0.03\\

    LipMIP & 12.13	&16.25&	102.64	&1,000.05*&	456.89&	1,000.12*
    &9.54&	1,000.08*	&57.88	&1,000.17*	&628.91&	1,000.40*\\
    
    LipSDP & 21.49&	11.01	&27.27	&103.04&	-&	-
    &11.03&	729.92&	-	&-&	-&	-\\
    
    LipBaB & 12.13&	2.92&	30.59&	63.16&	73.71&	60.98
    &5.39&	61.62	&15.34	&61.43	&36.40	&81.82\\
    
    RecurJac & 12.38&	17.40&	20.25	&17.07	&47.23&	16.66
   & 5.02&	16.21&	9.02&	16.73	&38.19&	16.38\\
   
    \midrule
   
    Ours (w/o BaB) & 12.28	&6.95	&17.45&	6.42&	35.66	&6.67
    &4.69	&7.29	&8.41&	7.40&	30.28	&7.29\\
    
    Ours (w/ BaB) & 12.13&	8.13	&16.30&	13.57&	28.69	&60.10
   & 4.69	&7.59&	8.19	&52.29	&28.6&	60.1\\
    
    \bottomrule
    \end{tabular}
    }
    \label{tab:synthetic_data}
\end{table}
\vspace{-10pt}

\subsection{Comparison on a Synthetic Dataset}
\label{eq:exp_synthetic_data}

We follow \citet{jordan2020exactly} and train several small models on a synthetic dataset. We compare different methods on models with varying width as shown in Table~\ref{tab:synthetic_data}, and we also show results with varying depth and $\epsilon$ in Appendix~\ref{apd:synthetic}. Some of these models are already large for LipMIP and we set a time limit of 1000 seconds. For our BaB, we stop further branching after 60 seconds.

Without BaB, our method is very efficient while it already outperforms the baselines on tightness except for LipMIP on the smallest MLP. Compared to LipMIP, for the smallest MLP on which LipMIP does not timeout, our result without BaB is very close  to but larger than their exact result. This smallest model serves as a sanity check to validate that our results should be no smaller than LipMIP's exact results. For other relatively larger models, LipMIP cannot produce exact results within the timeout and our result is much tighter than the upper bounds by LipMIP after 1000 seconds. And LipSDP fails to yield results within the timeout on many models. Although LipBaB has BaB to tighten the bounds, their results are still loose within the time budget, since using loose interval bounds in their BaB is inefficient. RecurJac is a relatively efficient baseline but still returns looser bounds compared to our method with tight linear relaxation. Our BaB further improves the bounds and consistently produces tightest results within a reasonable time budget.

\subsection{Comparison on Image Datasets}

\begin{table*}[ht]
    \centering
    \caption{Average local Lipschitz constant values and runtime (seconds) on MNIST. 
    The Lipschitz constants are evaluated on the first 100 examples in the test set.
    We set a timeout value of 120s for LipMIP and 60s for BaB. ``*'' denotes that we report the upper bound LipMIP returns at timeout, and ``-'' means LipMIP and LipBaB cannot return any valid upper bound at timeout on the CNN model. }    
    \adjustbox{max width=.55\textwidth}{
    \begin{tabular}{ccccccccc|cccc}
    \toprule
    \multirow{2}{*}{Method} & \multicolumn{2}{c}{3-layer MLP} & \multicolumn{2}{c}{CNN-2C2F}\\
    & Value & Runtime & Value & Runtime\\
    \midrule
    NaiveUB & 3,257.16	& 0.00 & 80,239.62 & 	0.00 \\
    LipMIP & 14,218.99* &	120.51 & - & - &\\
    LipBaB & 947.69	& 62.77 &  - & -\\
    RecurJac & 1,091.31	& 0.22  & 12,514.55 & 115.43\\
    \midrule
    Ours (w/o BaB) &   688.15 & 4.95 & 5,473.03 &	8.21\\
    Ours &  397.25 & 52.23  & 5,458.84	& 60.04\\
    \bottomrule
    \end{tabular}
    }
    \label{tab:mnist}
\end{table*}

\vspace{-10pt}
\begin{table*}[ht]
    \centering
    \caption{Average local Lipschitz constant values and runtime (seconds) on CIFAR-10 and TinyImageNet. The Lipschitz constants are evaluated on the first 100 examples in the test set. RecurJac's original implementation cannot handle large models here, and its results are obtained by supporting its relaxation in our implementation.}    
    \adjustbox{max width=.6\textwidth}{
    \begin{tabular}{ccccccccccccccc}
    \toprule
    Dataset & \multicolumn{2}{c}{CIFAR-10} & \multicolumn{2}{c}{TinyImageNet}\\
    Model & CNN-2C2F & CNN-4C2F & CNN-2C2F & CNN-4C2F\\
    \midrule
    NaiveUB & 1707252.88 & 365293440.00 & 1512185.25 & 87148664.00\\
    RecurJac & 79275.44 & 12502332.00 & 24031.58 & 714189.12\\
    \hline 
    Value (ours) & 18638.14 & 1049447.88 & 4556.61 & 25096.26 \\
    Runtime (ours) & 11.17 & 29.79 & 95.42 & 137.04\\
    \bottomrule
    \end{tabular}
    }
    \label{tab:cifar_tinyimagenet}
\end{table*}

\vspace{-10pt}
\begin{table*}[ht]
\centering
\caption{Percentage of examples on which the predicted confidence for high income level is monotonically increasing ($\uparrow$) or decreasing ($\downarrow$) w.r.t. each feature, by Recurjac and our method respectively.}
\adjustbox{max width=.8\textwidth}{
\begin{tabular}{ccccccc}
\toprule
Method & Monotonicity & Age & Education num & Capital gain & Capital loss  & Hours-per-week\\
\midrule
\multirow{2}{*}{RecurJac} & $\uparrow$ & 32\% & 55\% & 0\% & 4\% & 95\%\\
& $\downarrow$ & 0\% & 0\% & 0\% & 0\% & 0\% \\
\midrule
\multirow{2}{*}{Ours} & $\uparrow$ & 40\% & 58\% & 5\% & 7\% & 98\%\\
& $\downarrow$ & 0\% & 0\% & 0\% & 0\% & 0\%\\
\bottomrule
\end{tabular}
} 
\label{tab:adult}
\end{table*}

To further evaluate the scalability of different methods to relatively large models, we then conduct experiments on image datasets including MNIST~\citep{mnist}, CIFAR-10~\citep{cifar}, and TinyImagenet~\citep{le2015tiny}. We use a 3-layer MLP and a 4-layer CNN on MNIST, and we use 4-layer and 6-layer CNN models on CIFAR-10 and TinyImageNet. All the models here are too large for LipSDP. LipMIP and LipBaB can only handle the MLP model on MNIST, and they cannot return any valid upper bound for the CNN models within a reasonable time budget. RecurJac can handle the CNN model on MNIST, but other CNN models on CIFAR-10 and TinyImageNet are still too large for RecurJac after converted to equivalent MLP models. In contrast, our formulation with linear bound propagation allows us to utilize the efficient bound propagation for convolutional layers from \citet{xu2020automatic}. Since RecurJac is a special case under our formulation, we implement RecurJac's loose relaxation to obtain results for RecurJac on CNN models.

We show results on MNIST in Table~\ref{tab:mnist}.  Our results are much tighter compared to the baselines even if we do not enable BaB, and BaB can further tighten results. On CIFAR-10 and TinyImageNet, we do not enable BaB on to save time cost on relatively larger models, and we show results in Table~\ref{tab:cifar_tinyimagenet}. Our method can scale to these larger CNNs that previous works could not handle. And our results are  much tighter than NaiveUB and also RecurJac reimplemented in bound propagation. We present additional results on models with different random initialization in Appendix~\ref{ap:mnist_repeat}.

\subsection{An Application on Monotonicity Analysis}

We further use an application to demonstrate that our method can be used to verify the monotonicity of neural networks. Given a network, we aim to verify whether the network's output is a monotonic function w.r.t. each input feature. Monotonicity can be preferred properties in some real-world scenarios~\citep{sill1997monotonic,daniels2010monotone,liu2020certified,sivaraman2020counterexample}. In this experiment, we adopt the Adult dataset~\citep{blake1998uci}, with a binary classification task about predicting income level. There are only several continuous features, and we aim to check the monotonicity of income level w.r.t. age, education level, capital gain, capital loss, and hours per week. We train an MLP model and verify its monotonicity on the first 100 test examples. Signs of Clarke Jacobian can be used to check monotonicity~\citep{schaible1996generalized}, and we achieve it by checking the bounds of Clarke Jacobian.
For each continuous feature $j$, we set an input domain $\gX_j$ where only this particular feature can be varied between the minimum and maximum possible values in the dataset, while we keep other features fixed. And then we obtain bounds on the Clarke Jacobian of the output class for high income level, as $ \rmL_1 \leq \rmJ_1(\rvx) \leq \rmU_1~(\forall \rvx\in\gX_j)$. Then if $[\rmL_1]_{j}>0$, the predicted confidence on high income level is monotonically increasing w.r.t. feature $j$, and vice versa. 
We count the percentage of examples that satisfy each type of monotonicity respectively.  We show results in Table~\ref{tab:adult}. The predicted confidence on high income level is monotonically increasing w.r.t. age, education number, and hours per week on at least part of the examples, which is reasonable. On the other hand, the models learn little monotonicity on capital gain or loss features. Compared to RecurJac, our method verifies more monotonicity. We expect the gap to be larger when the models and datasets are larger.

%\vspace{-15pt}

\section{Conclusion}

In this paper, we propose an efficient framework to compute tight \rebuttal{$\ell_\infty$ local Lipschitz constants} for neural networks by upper bounding the norm of Clarke Jacobian, and we formulate the problem with linear bound propagation. 
We model the computation for the Clarke Jacobian as a backward computational graph and conduct linear bound propagation on the backward graph. 
We propose tight linear relaxation for nonlinearities in Clarke Jacobian with guarantees on the optimality. 
And we further tighten bounds with branch-and-bound when time budget allows.
Experiments show that our method efficiently produces much tighter results compared to existing relaxed methods and can scale to larger convolutional networks on which previous works cannot handle. We also use our method to analyze the monotonicity of neural networks as a potential application.

\paragraph{Limitations}
There are several limitations in this work and challenges for future works. It is still difficult for the proposed method to scale to deeper neural networks such as ResNet~\citep{he2016deep} with tens of convolutional layers on ImageNet~\citep{deng2009imagenet}. It is also a common challenge in neural network verification, where efficiency and the tightness of certified bounds can be very limited on deep models. Besides, we have only focused on $\ell_\infty$ local Lipschitz constants in this paper but not other norms. Linear bound propagation computes certified bounds for each neuron respectively and thus aligns better with $\ell_\infty$ norm. Bounds may become significantly looser if it is applied to other norms such as $\ell_2$ norm. Previous state-of-the-art works~\citep{jordan2020exactly,zhang2019recurjac,fazlyab2019efficient} on local Lipschitz constants did not consider both $\ell_\infty$ norm and $\ell_2$ norm simultaneously either.  Handling other norms is a challenge for future works. Moreover, in addition to verifying pre-trained models and computing local Lipschitz constants, we have not utilized the proposed method to train neural networks with stronger certified guarantees on Lipschitz constants or robustness, and it remains challenging for future works to better align verification and training.

\section*{Funding Disclosure}
This work is supported in part by NSF under  IIS-2008173, IIS-2048280 and by Army Research Laboratory under W911NF-20-2-0158. 
Huan Zhang is supported by a grant from the
Bosch Center for Artificial Intelligence.

\bibliography{reference,papers}
\bibliographystyle{icml2022}
\newpage
\section*{Checklist}

% %%% BEGIN INSTRUCTIONS %%%
% The checklist follows the references.  Please
% read the checklist guidelines carefully for information on how to answer these
% questions.  For each question, change the default \answerTODO{} to \answerYes{},
% \answerNo{}, or \answerNA{}.  You are strongly encouraged to include a {\bf
% justification to your answer}, either by referencing the appropriate section of
% your paper or providing a brief inline description.  For example:
% \begin{itemize}
%   \item Did you include the license to the code and datasets? \answerYes{See Section~\ref{gen_inst}.}
%   \item Did you include the license to the code and datasets? \answerNo{The code and the data are proprietary.}
%   \item Did you include the license to the code and datasets? \answerNA{}
% \end{itemize}
% Please do not modify the questions and only use the provided macros for your
% answers.  Note that the Checklist section does not count towards the page
% limit.  In your paper, please delete this instructions block and only keep the
% Checklist section heading above along with the questions/answers below.
% %%% END INSTRUCTIONS %%%

\begin{enumerate}

\item For all authors...
\begin{enumerate}
  \item Do the main claims made in the abstract and introduction accurately reflect the paper's contributions and scope?
    \answerYes{}
  \item Did you describe the limitations of your work?
    \answerYes{}
  \item Did you discuss any potential negative societal impacts of your work?
    \answerNA{Our work will enable potential users to better verify and understand the properties of neural networks, for more reliable and trustworthy deployment in the future. We did not find any negative social impact.}
  \item Have you read the ethics review guidelines and ensured that your paper conforms to them?
    \answerYes{}
\end{enumerate}

\item If you are including theoretical results...
\begin{enumerate}
\item Did you state the full set of assumptions of all theoretical results?
\answerYes{}
\item Did you include complete proofs of all theoretical results?
\answerYes{}
\end{enumerate}

\item If you ran experiments...
\begin{enumerate}
\item Did you include the code, data, and instructions needed to reproduce the main experimental results (either in the supplemental material or as a URL)?
\answerYes{}
  \item Did you specify all the training details (e.g., data splits, hyperparameters, how they were chosen)?
\answerYes{}
\item Did you report error bars (e.g., with respect to the random seed after running experiments multiple times)?
\answerNA{There is no randomness in our computation for local Lipschitz constants.}
\item Did you include the total amount of compute and the type of resources used (e.g., type of GPUs, internal cluster, or cloud provider)?
\answerYes{}
\end{enumerate}

\item If you are using existing assets (e.g., code, data, models) or curating/releasing new assets...
\begin{enumerate}
  \item If your work uses existing assets, did you cite the creators?
    \answerYes{}
  \item Did you mention the license of the assets?
    \answerYes{}
  \item Did you include any new assets either in the supplemental material or as a URL?
    \answerYes{}
  \item Did you discuss whether and how consent was obtained from people whose data you're using/curating?
    \answerYes{}
  \item Did you discuss whether the data you are using/curating contains personally identifiable information or offensive content?
    \answerYes{}
\end{enumerate}

\item If you used crowdsourcing or conducted research with human subjects...
\begin{enumerate}
  \item Did you include the full text of instructions given to participants and screenshots, if applicable?
    \answerNA{}
  \item Did you describe any potential participant risks, with links to Institutional Review Board (IRB) approvals, if applicable?
    \answerNA{}
  \item Did you include the estimated hourly wage paid to participants and the total amount spent on participant compensation?
    \answerNA{}
\end{enumerate}

\end{enumerate}

%%%%%%%%%%%%%%%%%%%%%%%%%%%%%%%%%%%%%%%%%%%%%%%%%%%%%%%%%%%%

\newpage
\appendix
\section{Additional Explanations}

\subsection{Linear Relaxation for ReLU}
\label{ap:relu}

\begin{figure}[ht]
\centering
\includegraphics[width=.5\textwidth]{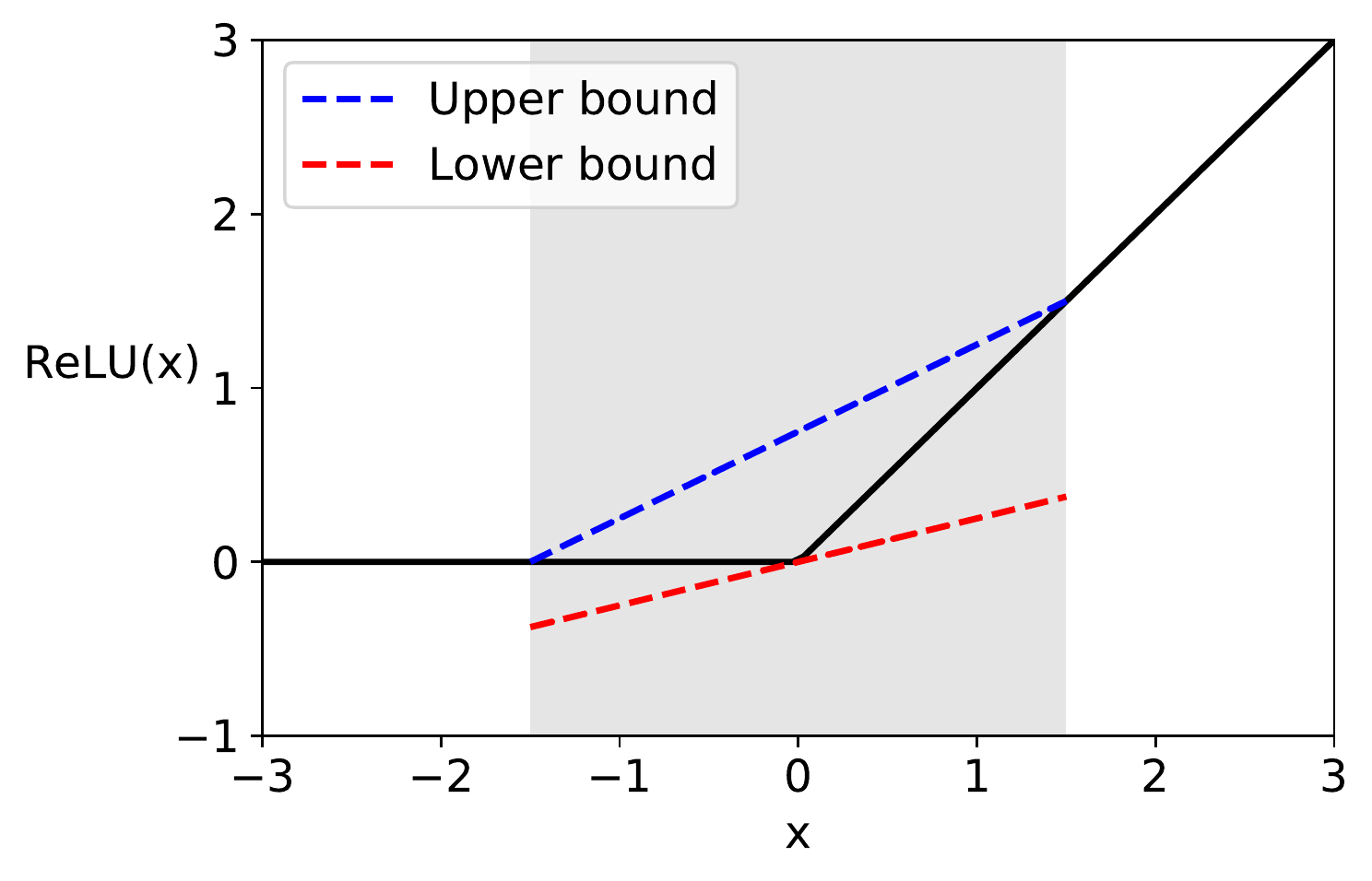}
\caption{Two lines can bound the ReLU activation when the sign of its input is unstable ($-1.5\leq x\leq 1.5 $ in this example) and are used for bound propagation with linear relaxation.}
\label{fig:relu}
\end{figure}

Recall that as introduced in Section~\ref{sec:bound_propagation},  given bounds of $z_i(\rvx)$ as $\rvl_i\leq z_i(\rvx)\leq \rvu_i$, we aim to relax an activation $\sigma(z_i(\rvx))$ as
$$
\forall \rvl_i \leq z_i(\rvx)\leq \rvu_i, \enskip 
\ul{\rvs}_i z_i(\rvx) + \ul{\rvt}_i \leq  \sigma(z_i(\rvx)) \leq \ol{\rvs}_i z_i(\rvx) + \ol{\rvt}_i.
$$
For every neuron $j$, if $ [\rvl_i]_j\!\geq\! 0$ or $[\rvu_i]_j\!\leq\! 0$, $\sigma(z_i(\rvx))\!=\!z_i(\rvx)$ or $ \sigma(z_i(\rvx))\!=\!0$ respectively is already linear, and then we can simply take 
$$  [\ul{\rvs}_i]_j [z_i(\rvx)]_j \!+\! [\ul{\rvt}_i]_j \!=\! [\ol{\rvs}_i]_j [z_i(\rvx)]_j \!+\! [\ol{\rvt}_i]_j\!=\!\sigma(z_i(\rvx)).$$ 
Otherwise when $ [\rvl_i]_j\!<\!0\!<\![\rvu_i]_j$,  the upper bound can be the line passing ReLU activation at $ [z_i(\rvx)]_j\!=\![\rvl_i]_j$ and  $ [z_i(\rvx)]_j\!=\![\rvu_i]_j$ respectively, i.e., 
$$ [\ol{\rvs}_i]_j[z_i(\rvx)]_j+[\ol{\rvt}_i]_j= \big( \frac{\sigma([\rvu_i]_j)-\sigma([\rvl_i]_j)}{[\rvu_i]_j-[\rvl_i]_j} \big) ( z_i(\rvx) - [\rvl_i]_j) + \sigma([\rvl_i]_j).$$
The lower bound can be any line with a slope between 0 and 1 and a bias of 0, i.e.,
$ 0\leq [\ul{\rvs}_i]_j\leq 1$, $ [\ul{\rvt}_i]_j=0$, where $[\ul{\rvs}_i]_j$ can be viewed as a parameter optimized with an objective of tightening output bounds~\citep{lyu2019fastened,xu2020fast,ryou2021scalable}. 
We illustrate the linear relaxation in Figure~\ref{fig:relu}.

\subsection{Linear Relaxation for Absolute Value Function}
\label{ap:abs}

\begin{figure}[ht]
\centering
\includegraphics[width=.5\textwidth]{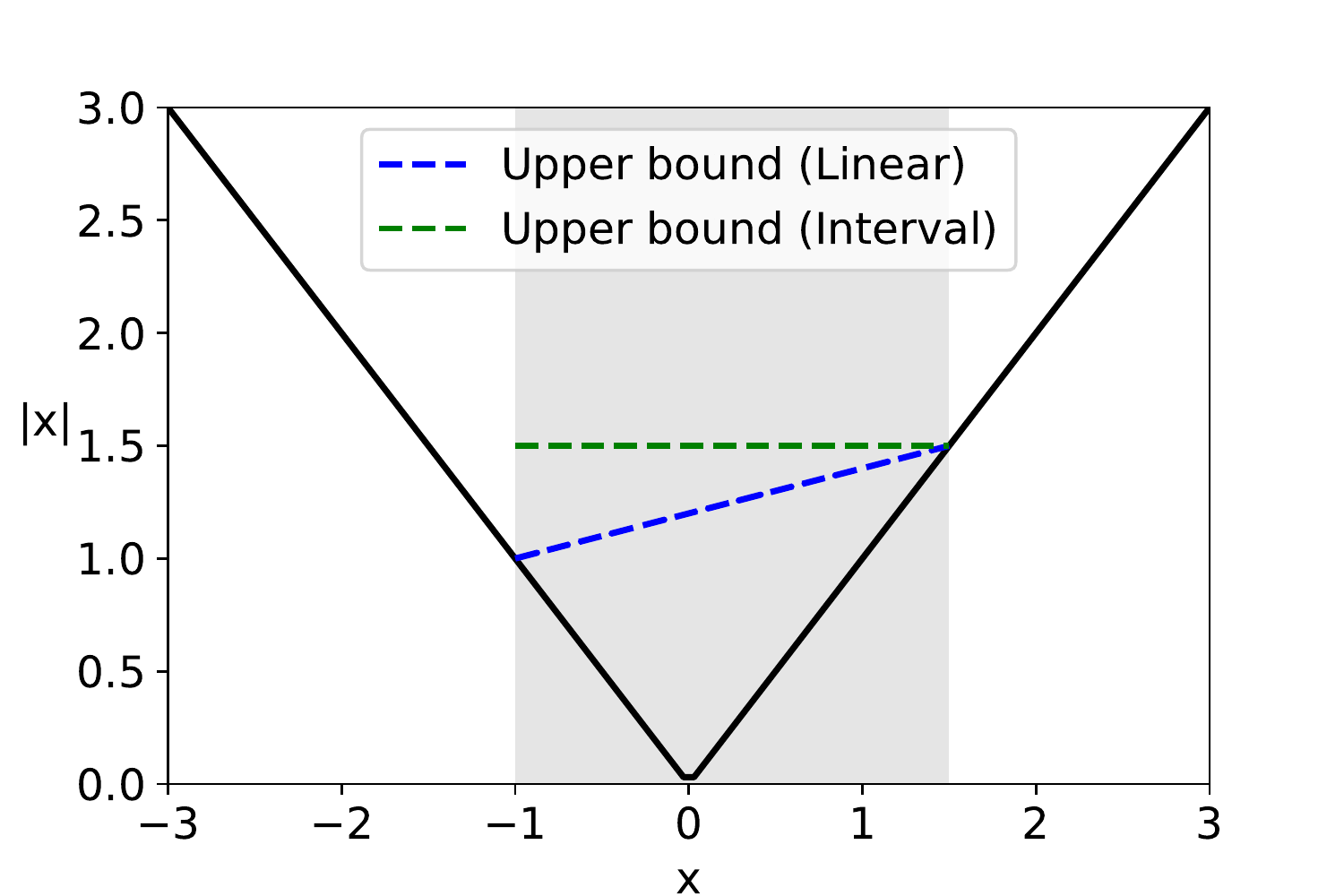}
\caption{Given the bound on the input to an absolute value function ($-1\leq x\leq 1.5$ in this example), a line can upper bound the function. We take the blue one with slope $\frac{1}{3}$, while \citet{zhang2019recurjac} would take the green one with slope 0, which is a looser upper bound.}
\label{fig:abs}
\end{figure}

In Figure~\ref{fig:abs}, we illustrate the linear relaxation for the absolute value function. 

\subsection{Applicability to Other Activations}

Although we focus on ReLU networks and $\ell_\infty$ local Lipschitz constants in this paper, our method can also be applied to networks with other activations.
When the activation is not ReLU, in Proposition~\ref{prop:grad_bound}, we can still obtain the range of $ [\Delta_i(\rvx)]_{jj}$ given pre-activation bounds $[\rvl_i]_j$ and $[\rvu_i]_j $. Suppose the Clarke gradient satisfies $ [\Delta_i(\rvx)]_{jj} \subseteq [l,u]$. 
Then 
$$ [\rmJ_{i+1}(\rvx)\Delta_i(\rvx)]_j\leq \max\{ [\rmJ_{i+1}(\rvx)]_j\cdot l, [\rmJ_{i+1}(\rvx)]_j\cdot u\}\coloneqq \tilde{\sigma}([\rmJ_{i+1}(\rvx)]_j), $$
and 
\begin{align*}
&[\rmJ_{i+1}(\rvx)\Delta_i(\rvx)]_j\\
\geq~& \min\{ [\rmJ_{i+1}(\rvx)]_j\cdot l, [\rmJ_{i+1}(\rvx)]_j\cdot u\}\\
=~ & -\max \{ -[\rmJ_{i+1}(\rvx)]_j\cdot l, -[\rmJ_{i+1}(\rvx)]_j\cdot u\} \\
=~ & -\tilde{\sigma}(-[\rmJ_{i+1}(\rvx)]_j),
\end{align*}
where $ \tilde{\sigma}(\cdot)$ is a Leaky-ReLU-like function.
Thereby, in Proposition~\ref{prop:relax_clarke_grad}, we can replace ReLU $\sigma(\cdot)$ with $ \tilde{\sigma}(\cdot)$ instead,
and the proposition will hold for non-ReLU activations in the network.

\section{Additional Experiments}

\subsection{Additional Experiments on Synthetic Data}
\label{apd:synthetic}

\begin{table}[ht]
\centering
\caption{
Local Lipschitz constant values and runtime (seconds) on MLP models with varying depth on synthetic data. Settings other than the models are the same as those for experiments in  Table~\ref{tab:synthetic_data}.
}  
\adjustbox{max width=.70\textwidth}{
    \begin{tabular}{c|cccccc}
    \toprule
    \multirow{3}{*}{Method} & \multicolumn{6}{c}{MLP with width=32} \\
    & \multicolumn{2}{c}{Depth=2} & \multicolumn{2}{c}{Depth=4} & \multicolumn{2}{c}{Depth=8} \\
    & Value & Runtime &  Value & Runtime &  Value & Runtime \\
    \midrule
    
    NaiveUB & 29.25	& 0.00 &	36.99	& 0.00	& 180.87	&0.00\\

    LipMIP & 15.42	&6.78	&12.32	&381.62	&18,938.35&	1,000.11*\\
    
    LipSDP & 25.88	&9.97	&24.57	&28.55	&49.00	&1,826.92\\
    
    LipBaB & 15.42	&5.87	&13.74&	60.10	&963.83	&60.67\\
    
    RecurJac & 15.45&	17.19	&13.95	&23.64	&195.16	&42.09\\
   
    \midrule
   
    Ours (w/o BaB) &16.00&	2.27	&13.24	&9.02	&35.64&	22.11 \\
    
    Ours (w/ BaB) & 15.42	&4.89	&12.36&	15.92	&30.32	&60.03\\
    
    \bottomrule
    \end{tabular}
    }
    \label{tab:synthetic_data_depth}
\end{table}

\begin{table}[ht]
\centering
\caption{
Local Lipschitz constant values and runtime (seconds) on an 3-layer MLP model with a width of 64, for input domains with varying radii $\eps$.
Settings other than the model and $\eps$ are the same as those for experiments in Table~\ref{tab:synthetic_data}.
}  
\adjustbox{max width=.80\textwidth}{
    \begin{tabular}{c|cccccccc}
    \toprule
    \multirow{3}{*}{Method} & \multicolumn{8}{c}{3-layer MLP with width=64} \\
    & \multicolumn{2}{c}{$\eps=0.01$} & \multicolumn{2}{c}{$\eps=0.05$} & \multicolumn{2}{c}{$\eps=0.1$} & \multicolumn{2}{c}{$\eps=0.2$} \\
    & Value & Runtime &  Value & Runtime &  Value & Runtime & Value & Runtime\\
    \midrule
    
    NaiveUB & 33.02	&0.01&	33.02	&0.01	&33.02	&0.01	&33.02	&0.01\\
   
    LipMIP & 15.49	&7.75&	15.71&	403.93	&102.64	&1,000.05* & 244.47	&1,000.05*\\
    
    LipSDP & 27.27 & 93.35 & 27.27 & 93.35 & 27.27 & 93.35 & 27.27 & 93.35\\
    
    LipBaB & 15.49&	2.66	&17.13	&60.44	&30.59	&63.16	&56.03&	60.26\\
    
    RecurJac & 15.49&	19.27	&16.17&	18.39	&20.25	&17.07&	65.28	&18.09\\
   
    \midrule
   
    Ours (w/o BaB) & 15.49	&4.32	&16.00&	5.41	& 17.45	&6.42&	40.34&	5.81\\
    
    Ours (w/ BaB) & 15.49	&5.46&	15.82&	13.54	&16.30&	13.57&32.70&	60.10\\
    
    \bottomrule
    \end{tabular}
    }
    \label{tab:synthetic_data_eps}
\end{table}

In this section, we show additional empirical results on the synthetic dataset. Settings are mostly similar to those for experiments in Section~\ref{eq:exp_synthetic_data}. 
In Table~\ref{tab:synthetic_data_depth}, we show results on models with varying depth. For deeper models, our method outperforms previous works with larger gaps.
And in Table~\ref{tab:synthetic_data_eps}, we show results on varying radii $\eps$ for the input domain. Our method outperforms previous methods for computing local Lipschitz constants (baselines except for NaiveUB and LipSDP) with larger gaps when the input domain has larger radii. But since local Lipschitz constants aim to analyze the properties of the network within a small local region, the input domain should not be too large, otherwise it is essentially no longer local.

\subsection{Repeated Experiments on MNIST}
\label{ap:mnist_repeat}

On MNIST, we show experiments where we train each of 3-layer MLP and CNN-2C2F with 5 different random initialization respectively, and other settings remain the same as those in Table~\ref{tab:mnist}. We compute local Lipschitz constants for these models, and we report the mean and standard deviations for the 5 runs in each setting respectively. We show the results in Table~\ref{tab:mnist_repeat}. Our improvement over the baselines is significant. 

\begin{table*}[ht]
    \centering
    \caption{Results on models with 5 different random initialization on MNIST. Other settings remain similar as those in  Table~\ref{tab:mnist}.}    
    \adjustbox{max width=.8\textwidth}{
    \begin{tabular}{ccccccccc|cccc}
    \toprule
    \multirow{2}{*}{Method} & \multicolumn{2}{c}{3-layer MLP} & \multicolumn{2}{c}{CNN-2C2F}\\
    & Value & Runtime & Value & Runtime\\
    \midrule
    NaiveUB & 3218.17 $\pm$ 289.99	& 0.00 $\pm$  0.00 & 92108.52  $\pm$ 13962.82 & 	0.00 $\pm$  0.00\\
    LipMIP & 14053.27 $\pm$  264.03*	& 120.29 $\pm$ 0.11 & - & - &\\
    LipBaB & 	988.66 $\pm$  86.93 & 	63.01 $\pm$  0.63 &  - & -\\
    RecurJac & 1153.68  $\pm$ 97.87	& 0.38 $\pm$ 0.08  & 	11174.06  $\pm$ 1342.49	& 114.18 $\pm$  3.40\\
    \midrule
    Ours (w/o BaB) &  \textbf{734.40 $\pm$ 63.69} & 4.26 $\pm$ 0.37 & \textbf{4895.50 $\pm$ 517.39} &  7.89 $\pm$ 0.20 \\
    Ours &  \textbf{423.30  $\pm$  38.25 } & 	58.54  $\pm$  3.15 & 
    \textbf{4878.96  $\pm$  518.49}  & 60.05  $\pm$  0.01 \\
    \bottomrule
    \end{tabular}
    }
    \label{tab:mnist_repeat}
\end{table*}

\section{Experiment Details}
\label{apd:exp_details}

\subsection{Synthetic Datasets}
\label{apd:synthetic_dataset}

For experiments using synthetic data, we generate the datasets following LipMIP~\citep{jordan2020exactly} which generates a randomized classification dataset given the input dimension and number of classes. 
We use 10 classes. We generate a dataset for MLP models and CNN models respectively, where the input dimension is $16$ for MLP models, and $1\times 8\times 8$ for CNN models. 
For each dataset, we generate 7000 examples for training and 3000 examples for testing. 

\subsection{Model Structures}
\label{apd:model_structure}

\paragraph{Synthetic Data} 
For experiments on synthetic data, the numbers of hidden neurons in MLP models and the number of convolution filters are listed as ``width'' in the tables for the corresponding results. The numbers of layers are also mentioned in the tables. All the convolution kernels have a size of $3\times 3$, a stride of 1, and no padding. Each CNN model only has one fully-connected output layer for classification, and there is no hidden fully-connected layer.

\paragraph{MNIST}
For experiments on MNIST, each hidden layer in the 3-layer MLP model has 20 neurons; and for the CNN-2C2F model, there are 8 convolution filters in each of the two convolutional layers, where each convolutional filter has a kernel size of $4\times 4$, a stride of 1, and no padding, and there is a fully-connected hidden layer with 100 neurons. 

\paragraph{CIFAR-10}
For experiments on CIFAR-10, the CNN-2C2F model has 32 convolution filters in each of the two convolutional laeyrs, where each convolutional filter has a kernel size of $3\times 3$, a stride of 1, and no padding, and there is a fully-connected hidden layer with 256 neurons;
and the CNN-4C2F model has two additional convolutional layers with same hyperparameters.

\paragraph{TinyImageNet}
For experiments on TinyImageNet, the CNN-2C2F model has 32 convolution filters in each of the two convolutional laeyrs, where each convolutional filter has a kernel size of $3\times 3$, a stride of 2, and a padding size of 1, and there is a fully-connected hidden layer with 256 neurons;
and the CNN-4C2F model has two additional convolutional layers with same hyperparameters.

\paragraph{Monotonicity Analysis}
For the monotonicity analysis, we use a 4-layer MLP model where each hidden layer has 512 neurons.

\subsection{Model Training}
\label{apd:training}

We use the Adam optimizer to train the models.
For experiments on the synthetic data, we train each model for 10 epochs with a learning rate of $10^{-3}$.
For experiments on the image datasets, we train each model for 30 epochs with a learning rate of $5\times 10^{-4}$.
For experiments on the monotonicity analysis, we train the model for 10 epochs with a learning rate of $5\times 10^{-4}$.
Hyperparameters were not specifically tuned, as the focus on this paper is not on training models.

\subsection{Compute Resources}

All experiments are done on internal GPU servers, and each experiment only uses one single GPU. We use a NVIDIA RTX A6000 GPU for experiments on CIFAR-10 and TinyImageNet, and we use a NVIDIA GeForce RTX 2080 Ti GPU for experiments on other datasets.

\subsection{Existing Assets}

The implementation is partly based on the following open-source code repositories under BSD-3-Clause license:
\begin{itemize}
    \item \texttt{auto\_LiRPA} (\url{https://github.com/Verified-Intelligence/auto\_LiRPA});
    \item \texttt{alpha-beta-CROWN}\\
    (\url{https://github.com/Verified-Intelligence/alpha-beta-CROWN}).
\end{itemize}

The following datasets are used, with no license found:
\begin{itemize}
    \item MNIST~ (\url{http://yann.lecun.com/exdb/mnist});
    \item CIFAR-10~(\url{https://www.cs.toronto.edu/~kriz/cifar.html});
    \item TinyImageNet~(\url{http://cs231n.stanford.edu/tiny-imagenet-200.zip});
    \item Adult dataset~(\url{https://archive.ics.uci.edu/ml/datasets/adult}).
\end{itemize}

These resources are publicly available. We believe the data do not contain
personally identifiable information or offensive content.
\section{Proofs for Our Linear Relaxation}

\subsection{Proof for Proposition~\ref{prop:abs_norm}}
\label{ap:abs_norm}

\begin{proof}
For every $j\in[d]$, we have $ [\rmL_1]_j\leq [\rmJ_1(\rvx)]_j\leq [\rmU_1]_j~(\forall \rvx\in\gX)$.
In the special case with $[\rmL_1]_j =[\rmU_1]_j=[\rmJ_1(\rvx)]_j$, we take 
$$[\tilde{\rmA}_1]_j=0, \enskip
[\tilde{\rvc}_0]_j = -[\widetilde{\rmA}_1]_j[\rmL_1]_j+\big|[\rmL_1]_j\big|
= \big|[\rmJ_1(\rvx)]_j\big|, $$
and thereby 
$ |[\rmJ_1(\rvx)]_j|= [\widetilde{\rmA}_1]_j [\rmJ_1(\rvx)]_j + [\widetilde{\rvc}_0]_j$.

Otherwise, assuming $[\rmL_1]_j<[\rmU_1]_j$, we have $ 0\leq \frac{[\rmJ_1(\rvx)]_j-[\rmL_1]_j}{[\rmU_1]_j-[\rmL_1]_j} \leq 1$.
Meanwhile, since $ |\cdot| $ is a convex function, we have 
\begin{align}
\forall 0\leq t\leq 1, \enskip | t [\rmU_1]_j + (1-t) [\rmL_1]_j  |  \leq t | [\rmU_1]_j | + (1-t) | [\rmL_1]_j |.
\label{eq:proof_prop1_1}
\end{align}
By taking $t=\frac{[\rmJ_1(\rvx)]_j-[\rmL_1]_j}{[\rmU_1]_j-[\rmL_1]_j}$, for \eqref{eq:proof_prop1_1}, the left-hand-side becomes
$$| t [\rmU_1]_j + (1-t) [\rmL_1]_j  | = |[\rmJ_1(\rvx)]_j|,$$
and the right-hand-side becomes
$$ t | [\rmU_1]_j | + (1-t) | [\rmL_1]_j| = \frac{( |[\rmU_1]_j|-|[\rmL_1]_j|) \cdot [\rmJ_1(\rvx)]_j - [\rmL_1]_j\cdot |[\rmU_1]_j|+|[\rmL_1]_j| \cdot [\rmU_1]_j}{[\rmU_1]_j - [\rmL_1]_j},$$
and thereby we have 
$$ |[\rmJ_1(\rvx)]_j|\leq \frac{( |[\rmU_1]_j|-|[\rmL_1]_j|) \cdot [\rmJ_1(\rvx)]_j - [\rmL_1]_j\cdot |[\rmU_1]_j|+|[\rmL_1]_j| \cdot [\rmU_1]_j}{[\rmU_1]_j - [\rmL_1]_j}.$$
By taking $ [\widetilde{\rmA}_1]_j=
\frac{|[\rmU_1]_j|-|[\rmL_1]_j|}{[\rmU_1]_j-[\rmL_1]_j},$ and $[\widetilde{\rvc}_0]_j=-[\widetilde{\rmA}_1]_j[\rmL_1]_j+|[\rmL_1]_j|$, the right-hand-side can be simplified into 
$ [\widetilde{\rmA}_1]_j [\rmJ_1(\rvx)]_j + [\widetilde{\rvc}_0]_j$, 
and thereby $ |[\rmJ_1(\rvx)]_j|\leq [\widetilde{\rmA}_1]_j [\rmJ_1(\rvx)]_j + [\widetilde{\rvc}_0]_j$.

So far, we have $ |[\rmJ_1(\rvx)]_j|\leq [\widetilde{\rmA}_1]_j [\rmJ_1(\rvx)]_j + [\widetilde{\rvc}_0]_j$ hold for all $j\in[d]$. 
In Section~\ref{sec:norm}, we have assumed that we handle one row in the Jacobian at each time, and $ \rmJ_1(\rvx)\in\sR^{1\times d}$ can be viewed as a row vector.
We upper bound its norm as 
$$\|\rmJ_1(\rvx)\|_\infty= \sum_{j=1}^d | [\rmJ_1(\rvx)]_j | \leq \sum_{j=1}^d  ([\widetilde{\rmA}_1]_j [\rmJ_1(\rvx)]_j + [\widetilde{\rvc}_0]_j)  =  \widetilde{\rmA}_1 \rmJ_1(\rvx) + \widetilde{\rvc}_0.$$
\end{proof}

\subsection{Proof for Proposition~\ref{prop:grad_bound}}
\label{ap:grad_bound}

\begin{proof}
For the $j$-th neuron in layer $i(i\in[n-1])$, given $ [\Delta_i(\rvx)]_{jj}=[0,1]$, if $ [\rmJ_{i+1}(\rvx)]_j\geq 0$, we have 
$$ [\rmJ_{i+1}(\rvx)\Delta_i(\rvx)]_j = 
[\rmJ_{i+1}(\rvx)]_j [\Delta_i(\rvx)]_{jj} \geq 0, $$    
and if $[\rmJ_{i+1}(\rvx)]_j<0$, we have 
$$ [\rmJ_{i+1}(\rvx)\Delta_i(\rvx)]_j = 
[\rmJ_{i+1}(\rvx)]_j [\Delta_i(\rvx)]_{jj} \geq [\rmJ_{i+1}(\rvx)]_j, $$
and thus 
$$ [\rmJ_{i+1}(\rvx)\Delta_i(\rvx)]_j \geq \min\{ [\rmJ_{i+1}(\rvx)]_j, 0 \}.$$
Similarly, we can also obtain 
$ [\rmJ_{i+1}(\rvx)\Delta_i(\rvx)]_j \leq\max\{ [\rmJ_{i+1}(\rvx)]_j ,0 \}$. 
Recall that ReLU activation is $\sigma(x)=\max\{x,0\}$, and thus  $\max\{ [\rmJ_{i+1}(\rvx)]_j, 0 \}=\sigma([\rmJ_{i+1}(\rvx)]_j)$,
and $ \min\{ [\rmJ_{i+1}(\rvx)]_j, 0 \} = -\max\{ -[\rmJ_{i+1}(\rvx)]_j, 0 \} = -\sigma(-[\rmJ_{i+1}(\rvx)]_j)$.
\end{proof}

\subsection{Proof for Proposition~\ref{prop:relax_clarke_grad}}
\label{ap:relax_clarke_grad}
\begin{proof}
For the $j$-th neuron in layer $i(i\in[n-1])$, given $ [\rmL_{i+1}]_j\leq[\rmJ_{i+1}(\rvx)]_j\leq [\rmU_{i+1}]_j$, we have $ -[\rmU_{i+1}]_j \leq -[\rmJ_{i+1}(\rvx)]_j\leq -[\rmL_{i+1}]_j$. And by the convex relaxation of ReLU activation in many previous works~\citep{wong2018provable,singh2019abstract,zhang2018efficient}, we have 
$$ \sigma([\rmJ_{i+1}(\rvx)]_j) \leq \frac{ \sigma([\rmU_{i+1}]_j) - \sigma([\rmL_{i+1}]_j) }{ [\rmU_{i+1}]_j - [\rmL_{i+1}]_j } ( [\rmJ_{i+1}(\rvx)]_j - [\rmL_{i+1}]_j) + \sigma([\rmL_{i+1}]_j), $$
$$  \sigma(-[\rmJ_{i+1}(\rvx)]_j) \leq \frac{ \sigma(-[\rmL_{i+1}]_j) - \sigma(-[\rmU_{i+1}]_j) }{ [\rmL_{i+1}]_j - [\rmU_{i+1}]_j } ( [\rmJ_{i+1}(\rvx)]_j + [\rmL_{i+1}]_j) +  \sigma(-[\rmL_{i+1}]_j),$$
and 
$$  -\sigma(-[\rmJ_{i+1}(\rvx)]_j) \geq -\frac{ \sigma(-[\rmL_{i+1}]_j) - \sigma(-[\rmU_{i+1}]_j) }{ [\rmL_{i+1}]_j - [\rmU_{i+1}]_j } ( [\rmJ_{i+1}(\rvx)]_j + [\rmL_{i+1}]_j) -  \sigma(-[\rmL_{i+1}]_j).$$
By setting
\begin{align*}
[\widetilde{\ul{\rvs}}_{i+1}]_j= \frac{ -\sigma(-[\rmU_{i+1}]_j) + \sigma(-[\rmL_{i+1}]_j) }{ [\rmU_{i+1}]_j - [\rmL_{i+1}]_j },
\enskip [\widetilde{\ul{\rvt}}_{i+1}]_j= - [\widetilde{\ul{\rvs}}_{i+1}]_j [\rmL_{i+1}]_j -\sigma(-[\rmL_{i+1}]_j),\\
[\widetilde{\ol{\rvs}}_{i+1}]_j= \frac{ \sigma([\rmU_{i+1}]_j) - \sigma([\rmL_{i+1}]_j) }{ [\rmU_{i+1}]_j - [\rmL_{i+1}]_j },
\enskip [\widetilde{\ol{\rvt}}_{i+1}]_j= - [\widetilde{\ol{\rvs}}_{i+1}]_j [\rmL_{i+1}]_j + \sigma([\rmL_{i+1}]_j),
\end{align*}
we have 
$$  -\sigma(-[\rmJ_{i+1}(\rvx)]_j) \geq [\widetilde{\ul{\rvs}}_{i+1}]_j [\rmJ_{i+1}(\rvx)]_j+[\widetilde{\ul{\rvt}}_{i+1}]_j.$$
$$ \sigma([\rmJ_{i+1}(\rvx)]_j) \leq [\widetilde{\ol{\rvs}}_{i+1}]_j [\rmJ_{i+1}(\rvx)]_j +[\widetilde{\ol{\rvt}}_{i+1}]_j. $$
By further combining results from Proposition~\ref{prop:grad_bound}, we have 
$$ [\rmJ_{i+1}(\rvx)\Delta_i(\rvx)]_j \geq -\sigma(-[\rmJ_{i+1}(\rvx)]_j)  \geq 
[\widetilde{\ul{\rvs}}_{i+1}]_j [\rmJ_{i+1}(\rvx)]_j + [\widetilde{\ul{\rvt}}_{i+1}]_j,$$
$$[\rmJ_{i+1}(\rvx)\Delta_i(\rvx)]_j \leq  \sigma([\rmJ_{i+1}(\rvx)]_j) \leq [\widetilde{\ol{\rvs}}_{i+1}]_j [\rmJ_{i+1}(\rvx)]_j + [\widetilde{\ol{\rvt}}_{i+1}]_j.$$
\end{proof}

\subsection{Proof for Theorem~\ref{theorem:tightness}}
\label{ap:tightness}

\begin{proof}
For the $j$-th neuron in the $i$-th layer ($i\in[n-1]$),
a linear relaxation $ \ul{s}J+\ul{t} $ is valid lower bound if and only if 
\begin{equation}
\forall J\!\in\! \Big[[\rmL_{i+1}]_j, [\rmU_{i+1}]_j\Big], \forall [\Delta_i(\rvx)]_{jj}\!\in\![0,1],
\enskip \ul{s}J+\ul{t} \leq J \cdot [\Delta_i(\rvx)]_{jj}. 
\label{eq:proof_tightness_1}
\end{equation}

When $ [\rmU_{i+1}]_j  \geq [\rmL_{i+1}]_j \geq 0$, 
we have 
$$ [\widetilde{\ul{\rvs}}_{i+1}]_j= \frac{ -\sigma(-[\rmU_{i+1}]_j) + \sigma(-[\rmL_{i+1}]_j) }{ [\rmU_{i+1}]_j - [\rmL_{i+1}]_j }
=  \frac{ 0 + 0 }{ [\rmU_{i+1}]_j - [\rmL_{i+1}]_j } = 0, $$
$$ [\widetilde{\ul{\rvt}}_{i+1}]_j= - [\widetilde{\ul{\rvs}}_{i+1}]_j [\rmL_{i+1}]_j -\sigma(-[\rmL_{i+1}]_j) = 0.$$
Then suppose there exists some $\hat{J}([\rmL_{i+1}]_j\!<\!\hat{J}\!<\![\rmU_{i+1}]_j)$ such that  
$ \ul{s}\hat{J}+\ul{t}>[\widetilde{\ul{\rvs}}_{i+1}]_j\hat{J}+[\widetilde{\ul{\rvt}}_{i+1}]_j$,
we have 
$ \ul{s}\hat{J}+\ul{t} >0$.
However, by taking $ J=\hat{J}$ and $[\Delta_i(\rvx)]_{jj}=0$, $\ul{s}J+\ul{t}>0$ while $J\cdot [\Delta_i(\rvx)]_{jj}=0$, and thus the condition $\ul{s}J+\ul{t}\leq J\cdot [\Delta_i(\rvx)]_{jj}=0 $ is violated, and $ \ul{s}J+\ul{t}$ cannot be a linear relaxation for the lower bound.
And if $ [\rmL_{i+1}]_j  \leq [\rmU_{i+1}]_j \leq 0$, 
we have 
$$ [\widetilde{\ul{\rvs}}_{i+1}]_j= \frac{ -\sigma(-[\rmU_{i+1}]_j) + \sigma(-[\rmL_{i+1}]_j) }{ [\rmU_{i+1}]_j - [\rmL_{i+1}]_j }
= \frac{ -(-[\rmU_{i+1}]_j) + (-[\rmL_{i+1}]_j) }{ [\rmU_{i+1}]_j - [\rmL_{i+1}]_j } = 1,$$
$$ [\widetilde{\ul{\rvt}}_{i+1}]_j= - [\widetilde{\ul{\rvs}}_{i+1}]_j [\rmL_{i+1}]_j -\sigma(-[\rmL_{i+1}]_j) = 0.$$
Then suppose there exists some $\hat{J}([\rmL_{i+1}]_j\!<\!\hat{J}\!<\![\rmU_{i+1}]_j)$ such that  
$ \ul{s}\hat{J}+\ul{t}>[\widetilde{\ul{\rvs}}_{i+1}]_j\hat{J}+[\widetilde{\ul{\rvt}}_{i+1}]_j$,
we have 
$ \ul{s}\hat{J}+\ul{t} > \hat{J}$.
However, by taking $ J=\hat{J}$ and $[\Delta_i(\rvx)]_{jj}=1$, $\ul{s}J+\ul{t}>\hat{J}$ while $J\cdot [\Delta_i(\rvx)]_{jj}=\hat{J}$, and thus the condition $\ul{s}J+\ul{t}\leq J\cdot [\Delta_i(\rvx)]_{jj}=\hat{J} $ is violated, and $ \ul{s}J+\ul{t}$ cannot be a linear relaxation for the lower bound.

Next, we consider the remaining case when $[\rmL_{i+1}]_j<0<[\rmU_{i+1}]_j$.
We have 
$$ [\widetilde{\ul{\rvs}}_{i+1}]_j= \frac{ -\sigma(-[\rmU_{i+1}]_j) + \sigma(-[\rmL_{i+1}]_j) }{ [\rmU_{i+1}]_j - [\rmL_{i+1}]_j }
= \frac{ -[\rmL_{i+1}]_j }{ [\rmU_{i+1}]_j - [\rmL_{i+1}]_j },$$
$$ [\widetilde{\ul{\rvt}}_{i+1}]_j= - [\widetilde{\ul{\rvs}}_{i+1}]_j [\rmL_{i+1}]_j -\sigma(-[\rmL_{i+1}]_j) = 
\frac{ [\rmL_{i+1}]_j  [\rmU_{i+1}]_j }{[\rmU_{i+1}]_j  - [\rmL_{i+1}]_j }.$$
Then suppose there exists some $\hat{J}([\rmL_{i+1}]_j\!<\!\hat{J}\!<\![\rmU_{i+1}]_j)$ such that  
$ \ul{s}\hat{J}+\ul{t}>[\widetilde{\ul{\rvs}}_{i+1}]_j\hat{J}+[\widetilde{\ul{\rvt}}_{i+1}]_j$,
we denote $ H = \ul{s}\hat{J}+\ul{t} $, and then $ \ul{s}J+\ul{t}=\ul{s}(J-\hat{J})+H$.
We have
$$ H > \frac{ -[\rmL_{i+1}]_j }{ [\rmU_{i+1}]_j - [\rmL_{i+1}]_j }\hat{J}+ \frac{ [\rmL_{i+1}]_j  [\rmU_{i+1}]_j }{[\rmU_{i+1}]_j  - [\rmL_{i+1}]_j },$$
and 
$$ H- [\rmL_{i+1}]_j 
= \frac{-[\rmL_{i+1}]_j (\hat{J}-[\rmL_{i+1}]_j)}{[\rmU_{i+1}]_j -[\rmL_{i+1}]_j }>0 \implies H> [\rmL_{i+1}]_j.$$
According to conditions in \eqref{eq:proof_tightness_1},
we have 
$$ \forall J\in\{ [\rmL_{i+1}]_j,\hat{J}, [\rmU_{i+1}]_j \}, \enskip \forall [\Delta_i(\rvx)]_{jj}\in[0,1], \enskip \ul{s} (J - \hat{J}) + H \leq J [\Delta_i(\rvx)]_{jj}.$$
Then by taking $J=\hat{J}$ and $[\Delta_i(\rvx)]_{jj}=0$, we have $H\leq 0$;
by taking $J= [\rmL_{i+1}]_j$ and $[\Delta_i(\rvx)]_{jj}=1$, we have 
$$  \ul{s} ([\rmL_{i+1}]_j - \hat{J}) + H \leq [\rmL_{i+1}]_j,$$
and thus 
$$ \ul{s} \geq \frac{H-[\rmL_{i+1}]_j}{\hat{J}-[\rmL_{i+1}]_j} > 0.$$
And by taking $ J=[\rmU_{i+1}]_j$, we have 
\begin{align*}
&\ul{s} ([\rmU_{i+1}]_j - \hat{J}) + H\\ 
\geq~ & \frac{H-[\rmL_{i+1}]_j}{\hat{J}-[\rmL_{i+1}]_j}([\rmU_{i+1}]_j - \hat{J}) + H\\
=~ & \frac{-[\rmL_{i+1}]_j }{[\rmU_{i+1}]_j -[\rmL_{i+1}]_j }([\rmU_{i+1}]_j - \hat{J}) + H \\
>~ & \frac{-[\rmL_{i+1}]_j }{[\rmU_{i+1}]_j -[\rmL_{i+1}]_j }([\rmU_{i+1}]_j - \hat{J}) + [\rmL_{i+1}]_j\\
=~ & \frac{[\rmL_{i+1}]_j ( \hat{J}-[\rmL_{i+1}]_j )}{[\rmU_{i+1}]_j -[\rmL_{i+1}]_j}>0,
\end{align*}
which violates \eqref{eq:proof_tightness_1} when $[\Delta_i(\rvx)]_{jj}=0$. 

Thus, so far we have proved that if $ \ul{s}J+\ul{t}$ is a valid linear relaxation for the lower bound, $ \forall J~([\rmL_{i+1}]_j <J<[\rmU_{i+1}]_j)$, 
$ \ul{s}\hat{J}+\ul{t}\leq [\widetilde{\ul{\rvs}}_{i+1}]_j\hat{J}+[\widetilde{\ul{\rvt}}_{i+1}]_j$ must hold, i.e., it must be no tighter than the relaxation proposed by Proposition~\ref{prop:relax_clarke_grad}.
And then we must have $ \forall J~([\rmL_{i+1}]_j <J<[\rmU_{i+1}]_j)$, 
$ \ul{s}\hat{J}+\ul{t}< [\widetilde{\ul{\rvs}}_{i+1}]_j\hat{J}+[\widetilde{\ul{\rvt}}_{i+1}]_j$, unless $(\ul{s},\ul{t})= ([\widetilde{\ul{\rvs}}_{i+1}]_j,[\widetilde{\ul{\rvt}}_{i+1}]_j)$.
And thereby,
$$(\ul{s},\ul{t})\neq ([\widetilde{\ul{\rvs}}_{i+1}]_j,[\widetilde{\ul{\rvt}}_{i+1}]_j) \implies \forall [\rmL_{i+1}]_j\!<\!J\!<\![\rmU_{i+1}]_j,\enskip\ul{s}J+\ul{t}<[\widetilde{\ul{\rvs}}_{i+1}]_jJ+[\widetilde{\ul{\rvt}}_{i+1}]_j.$$
Similarly, for the linear relaxation $ \ol{s}J+\ol{t} $ on the upper bound satisfying, 
$$\forall J\!\in\! \Big[[\rmL_{i+1}]_j, [\rmU_{i+1}]_j\Big], \forall [\Delta_i(\rvx)]_{jj}\!\in\![0,1],
\enskip \ol{s}J+\ol{t} \geq J \cdot [\Delta_i(\rvx)]_{jj}, $$
we can also prove
$$(\ol{s},\ol{t})\neq ([\widetilde{\ol{\rvs}}_{i+1}]_j,[\widetilde{\ol{\rvt}}_{i+1}]_j) \implies \forall [\rmL_{i+1}]_j\!<\!J\!<\![\rmU_{i+1}]_j,\enskip\ol{s}J+\ol{t}>[\widetilde{\ol{\rvs}}_{i+1}]_jJ+[\widetilde{\ol{\rvt}}_{i+1}]_j.$$
\end{proof}

Thereby, we have shown the optimality of our linear relaxation, which is provably tighter than other relaxations such as interval bound-like relaxation in RecurJac~\citep{zhang2019recurjac}. Alternatively, if we directly adopt the relaxation for the multiplication of two variables proposed in \citet{shi2019robustness} for the multiplication of $[\rmJ_{i+1}(\rvx)]_j$ and $[\Delta_i(\rvx)]_{jj}$, it will produce a similarly loose relaxation as \citet{zhang2019recurjac}.    

\section{Connection with RecurJac}
\label{ap:recurjac}

In this section, we explain in more detail that RecurJac~\citep{zhang2019recurjac} is a special case under our framework using linear bound propagation but RecurJac has used relatively loose interval bounds in non-trivial cases as partly illustrated in Figure~\ref{fig:clarke}.
For the $j$-th neuron in the $i$-th layer,
in cases where $[\rmL_{i+1}]_j<0<[\rmU_{i+1}]_j$
but excluding the situation where 
$$i=n-1,\enskip
[\Delta_i]_{jj}=0 ~\text{or}~ [\Delta_i]_{jj}=1,$$
if we use following interval relaxation instead of our relaxation proposed in Section~\ref{sec:grad_relax}, 
%if our relaxation in Proposition~\ref{prop:relax_clarke_grad} is replaced by the following interval relaxation:
\begin{align}
[\rmL_{i+1}]_j [\rvu'_i]_j 
\leq [\rmJ_{i+1}(\rvx)\Delta_i(\rvx)]_j 
\leq [\rmU_{i+1}]_j [\rvu'_i]_j 
\quad \text{for}\enskip
[\rmL_{i+1}]_j<0<[\rmU_{i+1}]_j,
%\,[\Delta_{i}(\rvx)]_{jj}\in[0,1],
\label{eq:use_interval_relaxation}
\end{align}
then our framework without BaB computes equivalent results as RecurJac given same bounds on $\Delta_i(\rvx)$.
This is an interval relaxation because the relaxed lower bound and upper bound no longer depend on $ \rmJ_{i+1}(\rvx)$ and thereby constitute an interval, instead of linear functions w.r.t. $ \rmJ_{i+1}(\rvx)$.
In particular, for the case illustrated in Figure~\ref{fig:clarke} with $ [\Delta_i]_{jj}=[0,1]$, \eqref{eq:use_interval_relaxation} is equivalent to 
$$ [\rmL_{i+1}]_j \leq [\rmJ_{i+1}(\rvx)\Delta_i(\rvx)]_j 
\leq [\rmU_{i+1}]_j 
\quad \text{for}\enskip
[\rmL_{i+1}]_j<0<[\rmU_{i+1}]_j.$$
But the interval relaxation is also used in more trivial cases with $[\Delta_i(\rvx)]_{jj}=0$ or $[\Delta_i(\rvx)]_{jj}=1$ fixed except $i=n-1$, where $ [\rmJ_{i+1}(\rvx)\Delta_i(\rvx)]_j $ is naturally a linear function of $\rmJ_{i+1}(\rvx)$ and can allow linear bound propagation to continue and propagate the bounds to $\rmJ_{i+1}(\rvx)$.
Using interval bounds also loosen the bounds in these cases.

Next, we compare with ``Algorithm 1'' in \citet{zhang2019recurjac} and show the equivalence. 
For each layer $i\in[n]$, we use $\hat{\rmL}_i$ and $\hat{\rmU}_i$ to denote the lower bound and upper bound on the Clarke Jacobian $\rmJ_i(\rvx)$ computed by our framework with the alternative relaxation in \eqref{eq:use_interval_relaxation}.
And we use $\tilde{\rmL}_i$ and $\tilde{\rmU}_i$ to denote bounds computed by RecurJac, i.e., ``$\rmL^{(-l)}$'' and ``$\rmU^{(-l)}$'' in \citet{zhang2019recurjac} where $l=i$.
We aim to show that $\hat{\rmL}_i=\tilde{\rmL}_i$ and $\hat{\rmU}=\tilde{\rmU}_i$ hold for all $ i\in[n]$.   

\paragraph{Equivalence on the last layer}
First, as mentioned in Section~\ref{sec:clarke_jacobian}, we have $\rmJ_n(\rvx)=\rmW_n$ which does not depend on any relaxation, and thus $ \hat{\rmL}_n=\hat{\rmU}_n=\rmJ_n(\rvx)=\rmW_n$ remains unchanged even if we change to use RecurJac's relaxation.
Meanwhile, RecurJac also returns $\tilde{\rmL}_n=\tilde{\rmU}_n=\rmW_n$ (in \citet{zhang2019recurjac}'s ``Algorithm 1'', it is showed that ``$ \rmL^{(-l)}=\rmU^{(-l)}=\rmW^{(l)}$'' when ``$ l=H$'', where $H$ is used to denote the number of layers in \citet{zhang2019recurjac} equivalent to $n$ here).

\paragraph{Equivalence on the second last layer}
Second, following \eqref{eq:jacobian}, we have $ \rmJ_{n-1}(\rvx) = \rmJ_n(\rvx)\Delta_{n-1}(\rvx)\rmW_{n-1}$. 
Note that the interval relaxation in \eqref{eq:use_interval_relaxation} is not used for layer $n-1$.
For every neuron $j$, if $[\Delta_i(\rvx)]_{jj}=0$ or $[\Delta_i(\rvx)]_{jj}=1$ is fixed, 
linear bound propagation is straightforward by merging  $[\Delta_{n-1}(\rvx)]_{jj}$ and $ [\rmW_{n-1}]_{j,:} $ into new linear coefficients:
\begin{align*}
[\rmJ_n(\rvx)]_j[\Delta_{n-1}(\rvx)]_{jj}[\rmW_{n-1}]_{j,:}
=[\rmJ_n(\rvx)]_j([\Delta_{n-1}(\rvx)]_{jj}[\rmW_{n-1}]_{j,:}).
\end{align*}
For other cases with $[\Delta_i(\rvx)]_{jj}=[0,1]$, note that since $ \rmJ_n(\rvx)=\hat{\rmL}_n=\hat{\rmU}_n=\rmW_n$ is constant, this is a special case and at least one of $ [\hat{\rmL}_n]_j=[\rmW_n]_j\geq 0$ and $[\hat{\rmU}_n]_j=[\rmW_n]_j\leq 0$ holds.
Then the relaxation in \eqref{eq:trivial_relaxation} is used.
Following \eqref{eq:apply_relaxation}, when using bound propagation to bound $\rmJ_{n-1}(\rvx)$, we have 
\begin{align*}
&\rmJ_{n-1}(\rvx)\\
=~& \rmJ_n(\rvx)\Delta_{n-1}(\rvx)\rmW_{n-1} \\
\geq~& \sum_{\substack{
[\Delta_{n-1}(\rvx)]_{jj}=0\\
\text{or}~[\Delta_{n-1}(\rvx)]_{jj}=1
}} [\rmJ_n(\rvx)]_j([\Delta_{n-1}(\rvx)]_{jj}[\rmW_{n-1}]_{j,:})\\
& + \sum_{\substack{
[\Delta_{n-1}(\rvx)]_{jj}=[0,1],\\
[\rmL_{n}]_j\geq 0
}}  [\rmJ_n(\rvx)]_j [\rmW_{n-1}]_{j,:,-}
+ \sum_{\substack{
[\Delta_{n-1}(\rvx)]_{jj}=[0,1],\\
[\rmU_{n}]_j\leq 0
}} [\rmJ_n(\rvx)]_j [\rmW_{n-1}]_{j,:,+} \\
%%%%%%
=~& \sum_{\substack{
[\Delta_{n-1}(\rvx)]_{jj}=0\\
\text{or}~[\Delta_{n-1}(\rvx)]_{jj}=1
}} [\rmW_n]_j([\Delta_{n-1}(\rvx)]_{jj}[\rmW_{n-1}]_{j,:})\\
& + \sum_{\substack{
[\Delta_{n-1}(\rvx)]_{jj}=[0,1],\\
[\rmL_{n}]_j\geq 0
}}  [\rmW_n]_j [\rmW_{n-1}]_{j,:,-}
+ \sum_{\substack{
[\Delta_{n-1}(\rvx)]_{jj}=[0,1],\\
[\rmU_{n}]_j\leq 0
}} [\rmW_n]_j [\rmW_{n-1}]_{j,:,+}.
\end{align*}
It is easy to verify that the above bound is equivalent to 
\begin{align*}
\rmJ_{n-1}(\rvx)\geq
 ([\rmW_n]_+ \rvl_{n-1} + [\rmW_n]_- \rvu_{n-1})[\rmW_{n-1}]_+
+  ([\rmW_n]_+ \rvu_{n-1} + [\rmW_n]_- \rvl_{n-1})[\rmW_{n-1}]_-. 
\end{align*}
Similarly, we also have 
\begin{align*}
\rmJ_{n-1}(\rvx)\leq
 ([\rmW_n]_+ \rvu_{n-1} + [\rmW_n]_- \rvl_{n-1})[\rmW_{n-1}]_+
+  ([\rmW_n]_+ \rvl_{n-1} + [\rmW_n]_- \rvu_{n-1})[\rmW_{n-1}]_-.
\end{align*}
These bounds are equivalent to \citet{zhang2019recurjac}'s Eq. (10) and (11) for obtaining $\tilde{\rmL}_{n-1}$ and $\tilde{\rmU}_{n-1}$, and thus $\hat{\rmL}_{n-1}=\tilde{\rmL}_{n-1}$ and $\hat{\rmU}_{n-1}=\tilde{\rmU}_{n-1}$.

\paragraph{Equivalence on remaining layers} 
Next, we use mathematical induction to show that $ \hat{\rmL}_i=\tilde{\rmL}_i$ and $\hat{\rmU}_i=\tilde{\rmU}_i$ hold for $i=n-2,n-3,\cdots,1$.    
Suppose we have shown that $ \hat{\rmL}_i=\tilde{\rmL}_i$ and $\hat{\rmU}_i=\tilde{\rmU}_i$ hold for $k+1\leq i\leq n$, and we aim to show that  $ \hat{\rmL}_{k}=\tilde{\rmL}_k$ and $\hat{\rmU}_k=\tilde{\rmU}_k$. We focus on the lower bound first, and the upper bound can be similarly proved.
$ \hat{\rmL}_k $ is the lower bound of $  \rmJ_k(\rvx)=\rmJ_{k+1}(\rvx)\Delta_k(\rvx)\rmW_k$ computed by bound propagation as 
\begin{align}
&\rmJ_{k+1}(\rvx)\Delta_k(\rvx)\rmW_k\nonumber\\
\geq~&
  \sum_{\substack{
  ([\Delta_{k}(\rvx)]_{jj}=0~\text{or}~[\Delta_{k}(\rvx)]_{jj}=1)\\
  \text{and}~( [\rmL_{k+1}]_j\geq 0 ~\text{or}~ [\rmU_{k+1}]_j\leq 0)
  }} 
  [\rmJ_{k+1}(\rvx)]_j ([\Delta_k(\rvx)]_{jj}[\rmW_k]_{j,:})\label{eq:recurjac_proof_2}\\
& 
  + \sum_{\substack{
  [\Delta_{k}(\rvx)]_{jj}=[0,1],\\
  [\rmL_{k+1}]_j\geq 0
  }}  [\rmJ_{k+1}(\rvx)]_j [\rmW_{k}]_{j,:,-}\label{eq:recurjac_proof_3}\\
&
  + \sum_{\substack{
  [\Delta_{k}(\rvx)]_{jj}=[0,1],\\
  [\rmU_{k+1}]_j\leq 0
  }} [\rmJ_{k+1}(\rvx)]_j [\rmW_{n-1}]_{j,:,+}\label{eq:recurjac_proof_4}\\
& 
  + \sum_{\substack{
  [\rmL_{k+1}]_j< 0<[\rmU_{k+1}]_j
  }}  
  ([\rmL_{k+1}]_j [\rvu'_i]_j  [\rmW_{k}]_{j,:,+}+[\rmU_{k+1}]_j [\rvu'_i]_j [\rmW_{k}]_{j,:,-}),
  \label{eq:recurjac_proof_5}
\end{align}
where \eqref{eq:recurjac_proof_2} is a special case with fixed $ [\Delta_k(\rvx)]_{jj}$ and the sign of $[\rmJ_{k+1}(\rvx)]_j$ is also fixed, 
\eqref{eq:recurjac_proof_3} and \eqref{eq:recurjac_proof_4} are by \eqref{eq:trivial_relaxation},
 and \eqref{eq:recurjac_proof_5} is by the interval relaxation in \eqref{eq:use_interval_relaxation}.
By merging, \eqref{eq:recurjac_proof_2}, \eqref{eq:recurjac_proof_3} and \eqref{eq:recurjac_proof_4}, the bound can be further simplified into
\begin{align}
&\rmJ_{k+1}(\rvx)\Delta_k(\rvx)\rmW_k\nonumber\\
\geq~
&
  \sum_{[\rmL_{k+1}]_j\geq 0} [\rmJ_{k+1}(\rvx)]_j [\rvu'_k]_j [\rmW_{k}]_{j,:,-}
  + [\rmJ_{k+1}(\rvx)]_j [\rvl'_k]_j [\rmW_{k}]_{j,:,+}\\
&
  \sum_{[\rmU_{k+1}]_j\leq 0} [\rmJ_{k+1}(\rvx)]_j [\rvl'_k]_j [\rmW_{k}]_{j,:,-}
  + [\rmJ_{k+1}(\rvx)]_j [\rvu'_k]_j [\rmW_{k}]_{j,:,+}\\  
& 
  + \sum_{\substack{
  [\rmL_{k+1}]_j< 0<[\rmU_{k+1}]_j
  }}  
  ([\rmL_{k+1}]_j [\rvu'_i]_j  [\rmW_{k}]_{j,:,+}+[\rmU_{k+1}]_j [\rvu'_i]_j [\rmW_{k}]_{j,:,-}).
  \label{eq:recurjac_proof_5}
\end{align}
\eqref{eq:recurjac_proof_5} corresponds to \citet{zhang2019recurjac}'s ``(14)''.
And when using linear bound propagation for cases with $[\rmL_{k+1}]_j\geq 0$ or $[\rmU_{k+1}]_j\leq 0$, bounds are propagated to $\rmJ_{k+1}(\rvx)$ and then the linear coefficients are merged with weights of the next layer $ \rmW_{k+1}$, which corresponds to  \citet{zhang2019recurjac}'s ``(17)''. 
Thus we have so far showed the equivalence in computing the lower bounds, and the equivalence can also be similarly derived for the upper bounds.

\end{document}